\documentclass[10pt,journal,compsoc]{IEEEtran}
\usepackage{amsmath}
\usepackage{amssymb}
\usepackage{graphicx} 
\graphicspath{{FIGURES/}}
\usepackage{helvet} 
\usepackage{booktabs}
\usepackage{diagbox}

\usepackage[final]{changes}

\ifCLASSOPTIONcompsoc
  \usepackage[nocompress]{cite}
\else
  \usepackage{cite}
\fi

\ifCLASSINFOpdf
  
\else
 
\fi

\usepackage{algorithmic}

\usepackage{array}

\ifCLASSOPTIONcompsoc
 \usepackage[caption=false,font=footnotesize,labelfont=sf,textfont=sf]{subfig}
\else
 \usepackage[caption=false,font=footnotesize]{subfig}
\fi

\usepackage{stfloats}

\usepackage{url}

\begin{document}
%
\title{Physical Adversarial Examples for Person Detectors in Thermal Images Based on 3D Modeling }

\author{Xiaopei~Zhu,
        Siyuan~Huang,
        Zhanhao~Hu,
        Jianmin~Li,
        Jun~Zhu,~\IEEEmembership{Fellow,~IEEE,}
        and~Xiaolin~Hu,~\IEEEmembership{Senior~Member,~IEEE}
\IEEEcompsocitemizethanks{
\IEEEcompsocthanksitem X. Zhu, S. Huang, Z. Hu, J. Li, J. Zhu and X. Hu are with the Department of Computer Science and
Technology, Institute for Artificial Intelligence, BNRist, Tsinghua University, Beijing
100084, China. X. Hu is also with the Chinese Institute for Brain Research (CIBR), Beijing 100010, China.\protect\\
E-mail: \{zxpthu, siyuanhuang92, zhanhaohu.cs\}@gmail.com,\\
\{lijianmin, dcszj, xlhu\}@tsinghua.edu.cn

Corresponding author: Xiaopei Zhu
}
}

\markboth{SUBMISSION TO IEEE TRANSACTIONS JOURNAL}%
{Zhu \MakeLowercase{\textit{et al.}}: Physical Adversarial Examples for Person Detectors in Thermal Images Based on 3D Modeling}

\IEEEtitleabstractindextext{%
\begin{abstract}
Thermal Infrared detection is widely used in autonomous driving, medical AI, etc., but its security has only attracted attention recently. 
We propose infrared adversarial clothing designed to evade thermal person detectors in real-world scenarios. The design of the adversarial 
clothing is based on 3D modeling, which makes it easier to simulate multiangle scenes near the real world compared to 2D modeling. 
We optimized the black patch layout pattern of 3D clothing based on the adversarial example technique and made physical adversarial 
clothing using the aerogel. The idea is to paste a set of square aerogel patches, which display black squares in thermal images, 
in the inner side of clothing at specific locations with specific orientations. To enhance realism, we propose a method to build 
infrared 3D models with real infrared photos and develop texture maps for 3D models to simulate varied infrared characteristics over 
time and location. In physical attacks, we achieved an attack success rate of 80.11\% indoors and 76.85\% outdoors against YOLOv9. 
In contrast, randomly placed patches yielded much lower success rates (26.53\% indoors and 23.03\% outdoors). 
The adversarial clothing also showed good transferability to unknown detectors with an ensemble attack method, demonstrating the 
effectiveness of our approach.
\end{abstract}

}

\maketitle

\IEEEdisplaynontitleabstractindextext

\IEEEpeerreviewmaketitle

\IEEEraisesectionheading{\section{Introduction}\label{sec:introduction}}

\IEEEPARstart{D}{eep} learning models can be fooled by carefully designed inputs, which are called adversarial examples. Adversarial examples include digital and physical varieties.
The study of physical adversarial examples has great significance for the security 
of AI systems, because deep learning-based technologies such as face recognition and autonomous driving are now widely used in the real world.

Adversarial examples not only exist in the visible light domain \cite{conf/cvpr/ThysRG19,journals/corr/abs-1910-11099,hu2021naturalistic,conf/cvpr/EykholtEF0RXPKS18,hu2022cvpr} 
but also in the thermal infrared (infrared for short) domain \cite{zhu2021fooling,kim2022map,zhu2022cvpr}. In recent years, infrared 
imaging has been widely used in autonomous driving, medical diagnose, and body temperature measurement. 
Infrared imaging has unique advantages \cite{zhu2021fooling}. First, infrared images contain temperature 
information. Second, infrared imaging can work even if
there is occlusion. Third, infrared imaging does not depend on external light sources and can work at night. 
Research on physical infrared adversarial examples has important significance for the safety of infrared imaging systems
but attracted little attention.

\begin{figure}[htbp]
\centering
\includegraphics[width=1\columnwidth]{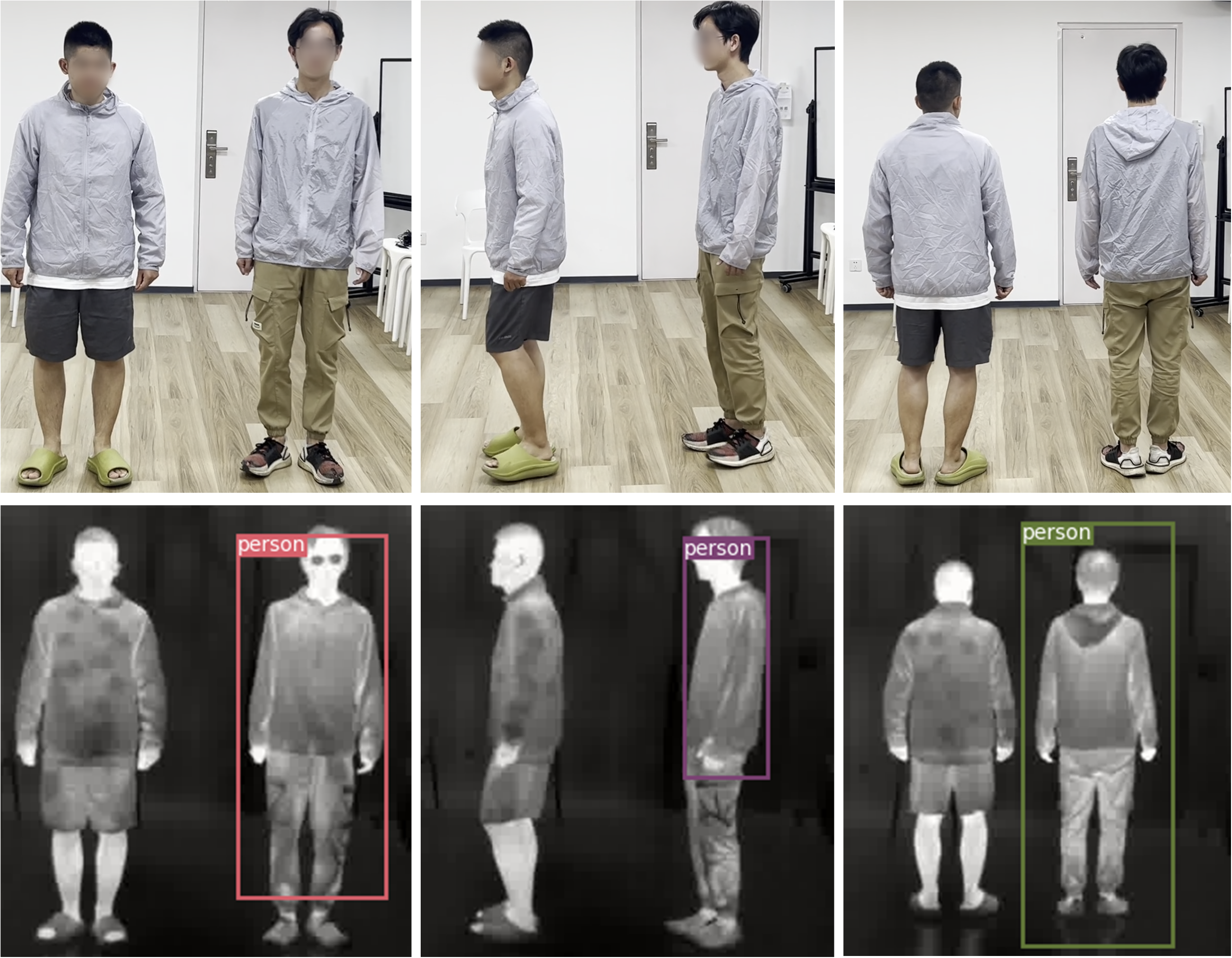} 
\caption{One example of infrared physical attacks. The adversarial clothing looks like ordinary clothing in the visible light view. The facial areas are blurred for privacy reasons.
The person wearing adversarial clothing is not detected by the detectors at multiple angles in infrared imaging, while the other person wearing ordinary clothes is detected (indicated by the bounding boxes). }
\label{one_example}
\end{figure} 

Many studies \cite{conf/cvpr/ThysRG19,journals/corr/abs-1910-11099,hu2021naturalistic,conf/cvpr/EykholtEF0RXPKS18,zhu2021fooling,kim2022map} about physical adversarial examples are based on the 2D patch. 
However, these studies have limitations. The 2D patch does not cover the entire 3D
surface of the object. For example, the adversarial T-shirt \cite{journals/corr/abs-1910-11099} can only perform attacks from the front of the human body and does not work from other angles (e.g., the side of the body).

Adversarial examples based on 3D modeling help to solve the above problem. With 3D modeling, we can design adversarial patterns from different angles. Compared with the 2D image, the 3D model is closer to real-world objects, which can
simulate the physical adversarial attacks more realistically.

In the visible light domain, there have been some physical adversarial studies \cite{conf/icml/AthalyeEIK18,conf/cvpr/HuangGZXYZL20,wang2020can,zhang2019camou} based on 3D modeling.
However, besides the difference in the modality (infrared and visible light) we are concerned with, there are other important differences between these studies and the present study. Specifically, the study in \cite{zhang2019camou} 
used nondifferentiable rendering and trained a neural network to simulate it. In contrast, we propose to use a differentiable render to make the optimization more accurate.
Different from the previous studies \cite{conf/cvpr/HuangGZXYZL20, wang2020can} where detectors were attacked at a narrow range (e.g., -50 to 50 degrees), we attack the detectors through the full range of 0-360 degrees.
Different from the classifier attack \cite{conf/icml/AthalyeEIK18}, we focus on the object detection attack, which is more challenging.

To the best of our knowledge, there is currently a lack of 3D model-based physical adversarial research in the
infrared domain.
Infrared 3D models are constructed based on the shape and temperature of the object. The pixel values of grayscale images reflect the temperature of the surface of the object.
Unlike the RGB 3D model, the pattern of the infrared 3D model is not easily ``printed'' in the physical world.

Recently, we designed a QR code-like clothing to attack person detectors from multiple angles\cite{zhu2022cvpr}, but that method is based on 2D pattern design and does not use 3D modeling of either a person or clothing.
That approach has several limitations. Due to the lack of the 3D model, there is a big gap between 
the 2D digital simulation and 3D physical implementation. 
Besides, the thermal insulation materials are pasted outside the clothing, which looks strange. In addition, the manufacturing process of the clothing is quite complex and requires a
lot of human power (7 persons over 15 days) and material costs.

In this study, we propose 3D model-based physical infrared adversarial clothing.
To simulate the infrared characteristics more realistically, we propose a method to build 
infrared 3D models with real infrared photos. Using above method, we build 3D digital models with real 
infrared characteristics for both the human body and clothing to minimize the gap between the digital 
simulation and physical implementation. We also build different texture maps for 3D clothing model to 
simulate changeable infrared clothing characteristics in different times and places. 
Then, we manufacture the adversarial clothes in the physical world.
Adversarial clothing looks similar to ordinary clothing in visible light imaging but can be invisible 
to object detectors in infrared imaging.
In addition, the proposed method requires much fewer materials and makes it easier to manufacture 
clothes than the QR code approach \cite{zhu2022cvpr}. Figure \ref{one_example} shows an example.
To the best of our knowledge, this study is the first to investigate infrared physical attacks based 
on 3D modeling. 


\section{Related Works}\label{sec2}

\subsection{2D adversarial attacks}\label{subsec2}

\added{2D adversarial attacks can be divided into digital attacks and physical attacks. 
Physical adversarial attacks have attracted significant attention in recent years due to their 
high practical value in real-world applications.}

\subsubsection{2D digital attacks}\label{subsubsec2}

Adversarial attacks originate from the vulnerability of deep neural networks (DNNs) revealed by Szegedy et al. \cite{journals/corr/SzegedyZSBEGF13}. Early adversarial attacks focused on the digital domain \cite{conf/sp/Carlini017,journals/corr/GoodfellowSS14,conf/iclr/KurakinGB17,conf/ijcai/XiaoLZHLS18,conf/aaai/LiuLFMZXT19}.
The input with carefully designed perturbations causes DNNs to make mistakes in the digital world. 2D digital attacks are usually classified into three types: gradient-based methods \cite{journals/corr/GoodfellowSS14,conf/iclr/MadryMSTV18,moosavi2016deepfool,conf/iclr/KurakinGB17}, \cite{bai2020adversarial,yuan2024adaptive}, optimization-based methods \cite{conf/sp/Carlini017,DBLP:conf/ccs/ChenZSYH17,journals/corr/SzegedyZSBEGF13}, \cite{yin2023generalizable,shi2022query}, and GAN-based
methods \cite{conf/aaai/LiuLFMZXT19,conf/ijcai/XiaoLZHLS18,conf/icb/DebZJ20}. Digital attacks 
assume that attackers can modify the input of the model, 
which is difficult to achieve in the real world.
However, digital attacks are the basis for physical attacks.

\subsubsection{2D physical attacks}

\added{Physical attacks are designed to attack DNNs in the real world. Most physical attacks focus on the visible light domain, and 
infrared physical attacks have only attracted attention in recent years.}

\added{In the visible-light domain, researchers typically design pixel-level RGB patterns and then print these patterns on paper, stickers, 
or clothing for physical display. For adversarial papers, Thys et al. \cite{conf/cvpr/ThysRG19} printed the adversarial example on 
a piece of paper, making the detector unable to identify the person holding the paper. Du et al. \cite{du2022physical} attached 
adversarial paper on car roofs and parking lines to attack an aerial imagery object detector. }

\added{For adversarial stickers, Eykholt et al. \cite{conf/cvpr/EykholtEF0RXPKS18} develop stickers with robust physical perturbations 
(RP2) to mask the stop sign at specific locations, making DNNs misclassify the stop sign as a speed limit sign. Tao et al. 
\cite{bai2021inconspicuous} designed stickers with perturbations that are nearly imperceptible to the human eye, 
causing DNNs to misclassify various signs. Wei et al. \cite{wei2022simultaneously} simultaneously optimized the positions 
and perturbations for 2D adversarial patch attacks and attacked face and traffic sign recognition models in the black-box 
setting. Wei et al. \cite{wei2022adversarial} proposed adversarial stickers to achieve black-box attacks in the physical world, 
which is a physically feasible and stealthy attack method. }

\added{For adversarial clothes, Xu et al. \cite{journals/corr/abs-1910-11099} proposed an adversarial 
T-shirt and used the thin plate spline (TPS) method to simulate clothing deformation. Wu et al. \cite{wu2020making} attacked 
YOLOv2 using wearable clothes in the physical world, and evaluated the performance of such attacks with different metrics. 
Hu et al. \cite{hu2021naturalistic} used GAN to generate adversarial examples, which made them more natural when printed 
on clothes. We \cite{hu2022cvpr} proposed the adversarial texture, which was printed on clothes to fool the visible light 
person detector at multiple angles.}


\added{In the infrared domain, researchers typically perform mathematical modeling of the infrared characteristics 
of basic components, then design algorithms to optimize the key parameters of these components, and finally conduct 
physical implementation based on the optimized parameters. Physical implementation methods can be divided into two 
categories: one based on heating or cooling elements, and the other based on insulating materials. For heating or 
cooling device-based methods, we \cite{zhu2021fooling } proposed an adversarial board based on small bulbs to fool 
the infrared pedestrian detector YOLOv3, and then we expand this work to attack the infrared and visible detectors 
simultaneously. After that, we \cite{zhu2023hiding} proposed the adversarial clothes based on carbon heaters to 
deceive infrared pedestrian detectors. Wei et al. \cite{wei2023hotcold} applied warming and cooling paste to 
hide from the infrared person detectors. Hu et al. \cite{hu2024AIB} proposed adversarial infrared blocks to attack infrared detectors 
in a black-box setting. After that, they \cite{hu2024AIC} proposed the adversarial infrared curves based on Bezier curve optimization 
to further enhance the efficiency in physical deployment. Tiliwalidi et al. \cite{tiliwalidi2025advgrid} proposed the AdvGrid to hide from the 
infrared pedestrian detectors using readily available materials and minimal setup. }

\added{For thermal insulation material-based methods, we \cite{zhu2022cvpr} proposed the adversarial ``QR codes'' 
texture and made the adversarial clothing to fool the infrared pedestrian detectors at multiple angles. 
Kim et al. \cite{kim2022map} proposed the adversarial board to attack infrared detector using materials 
with different emissivity. After that, they \cite{kim2023multispectral} propose the adversarial clothes based on low-E films to 
conduct cross-modal attacks. Wei et al. \cite{wei2023physically} proposed infrared adversarial patches with 
learnable shapes and locations. Wei et al. \cite{wei2023unified} proposed unified adversarial patches for 
cross-modal attacks in the physical world. }

\subsection{3D adversarial attacks}
\added{2D adversarial attacks usually exhibit effectiveness only within a narrow angular range 
(e.g., approximately -50$^\circ$ to 50$^\circ$), whereas 3D adversarial attacks can cover a 
full 0–360$^\circ$ viewpoint range. Currently, 3D adversarial attacks mainly focus on the visible 
and LiDAR domains, while 3D infrared attacks have not been fully explored.}

\subsubsection{3D digital attacks}

\added{In the visible-light field, 
Liu et al. \cite{liu2018beyond} optimized the perturbations by changing conditions such as lighting so that the 3D 
shoe model was incorrectly classified when only the lighting changed. Xiao et al. \cite{xiao2019meshadv} optimized 
the rendering to generate the 3D adversarial car model in the digital world. Wang et al. \cite{wang2020can} proposed 
the 3D adversarial logo to fool the person detector in the digital world. Wang et al. \cite{wang2022fca} proposed a full-coverage 
vehicle camouflage for multi-view physical adversarial attack. Suryanto et al. \cite{suryanto2022dta} proposed a differentiable 
transformation network (DTN) to simulate the adversarial car attack realistically in the digital world. }

\added{In the LiDAR field, Xiang et al. \cite{xiang2019generating} generated 3D point cloud perturbations by 
shifting, reshaping, or repositioning the position of the point cloud so that the DNNs misidentified the point 
cloud of the bottle as various objects. Tu et al. \cite{conf/cvpr/TuRMLYDCU20} optimized a 3D mesh of a specific 
shape and then placed the mesh on the roof of a 3D car model, which fooled the 3D LiDAR detector.}

\subsubsection{3D physical attacks}


\added{In the visible domain, 
Athalye et al. \cite{conf/icml/AthalyeEIK18} 3D-printed an adversarial turtle that was misclassified by a DNN as a 
rifle. Huang et al. \cite{conf/cvpr/HuangGZXYZL20} proposed Universal Physical Camouflage (UPC), where they 
simulated an adversarial attack using a 3D model and then attacked a detector in the physical world. 
Wang et al. \cite{wang2021dual} proposed a dual attention suppression (DAS) method to optimize adversarial stickers on 3D 
vehicle surfaces, and Li et al. \cite{li2024flexible} introduced a flexible physical camouflage generation method based on a 
differential optimization approach.}

\added{In the LiDAR field, Cao et al. \cite{cao2019adversarial} proposed an adversarial sensor attack on LiDAR-based perception in 
autonomous driving. Subsequently, Jin et al. \cite{jin2023pla} developed a physical laser attack against LiDAR-based 3D object 
detection in autonomous driving. Additionally, Cao et al. \cite{cao2021invisible} generated interfering obstacles for LiDAR by 
3D-printing adversarial geometric shapes and optimizing their forms to achieve adversarial effects.}

\subsection{Infrared stealth materials}

Traditional infrared stealth materials primarily include iron composites \cite{wu2022infrared}, zinc composites \cite{yang2020compatible}, aluminum composites \cite{fan2020microwave}, etc. They have low infrared emissivity due to their good electrical conductivity. Among them, aluminum has the lowest cost
and is widely used. However, metal composites are susceptible to surface topography. Therefore, some studies have turned to addressing the limitations of metal composites triggered by semiconductor materials. Shang et
al. \cite{shang2019microstructure} observed that silica composite aerogels could exhibit good thermal insulation stability at room temperature. Wang et al. \cite{wang2021polysiloxane} 
proposed bonding the aerogel with polysiloxane to enhance its thermal insulation ability. These studies show that the aerogel has excellent thermal insulation properties.

\begin{figure*}[htbp]
\centering
\includegraphics[scale=0.8]{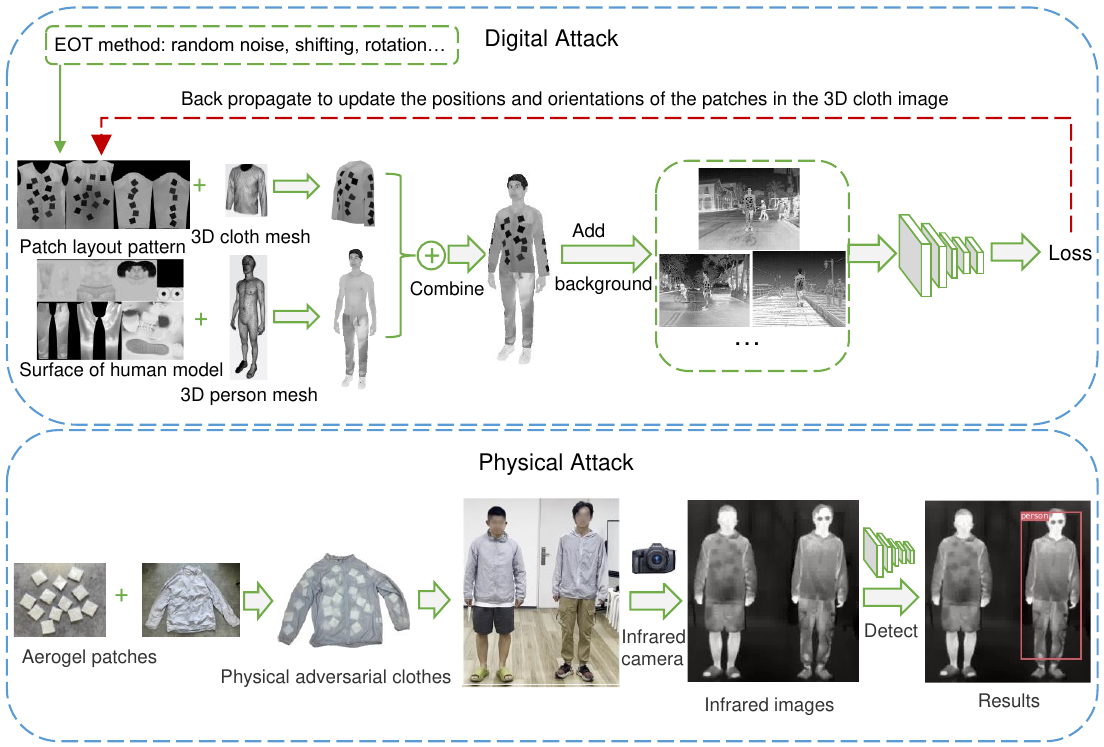} 
\caption{Pipeline of the proposed method. Top: optimization in the digital world. Bottom: physical test process. The facial areas
are blurred for privacy reasons.}
\label{main_process}
\end{figure*} 

\section{Methods} \label{sec:methods}
We first propose a method to build infrared 3D models with real infrared photos. Then, we build 
the 3D infrared human body model and adversarial clothing model. To make 3D modeling more realistic and adaptable, 
we simulate infrared clothing characteristics in different times and places. After that, 
we design the adversarial pattern in 3D clothing and define the optimization loss function.
Finally, we introduce the physical implementation of adversarial clothing. 
\added{Figure \ref{main_process} shows the pipeline of the proposed method.}

\subsection{Building 3D infrared models with real infrared photos} \label{sec:infared_model}
A naive infrared modeling method is to directly convert 3D RGB models to 
grayscale models. Although infrared images are also grayscale images, there is a big domain gap between 
the rendered 3D models and the real infrared images due to the huge difference in imaging mechanisms 
between RGB cameras and infrared cameras.

\begin{figure}[htbp]
\centering
\includegraphics[width=1\columnwidth]{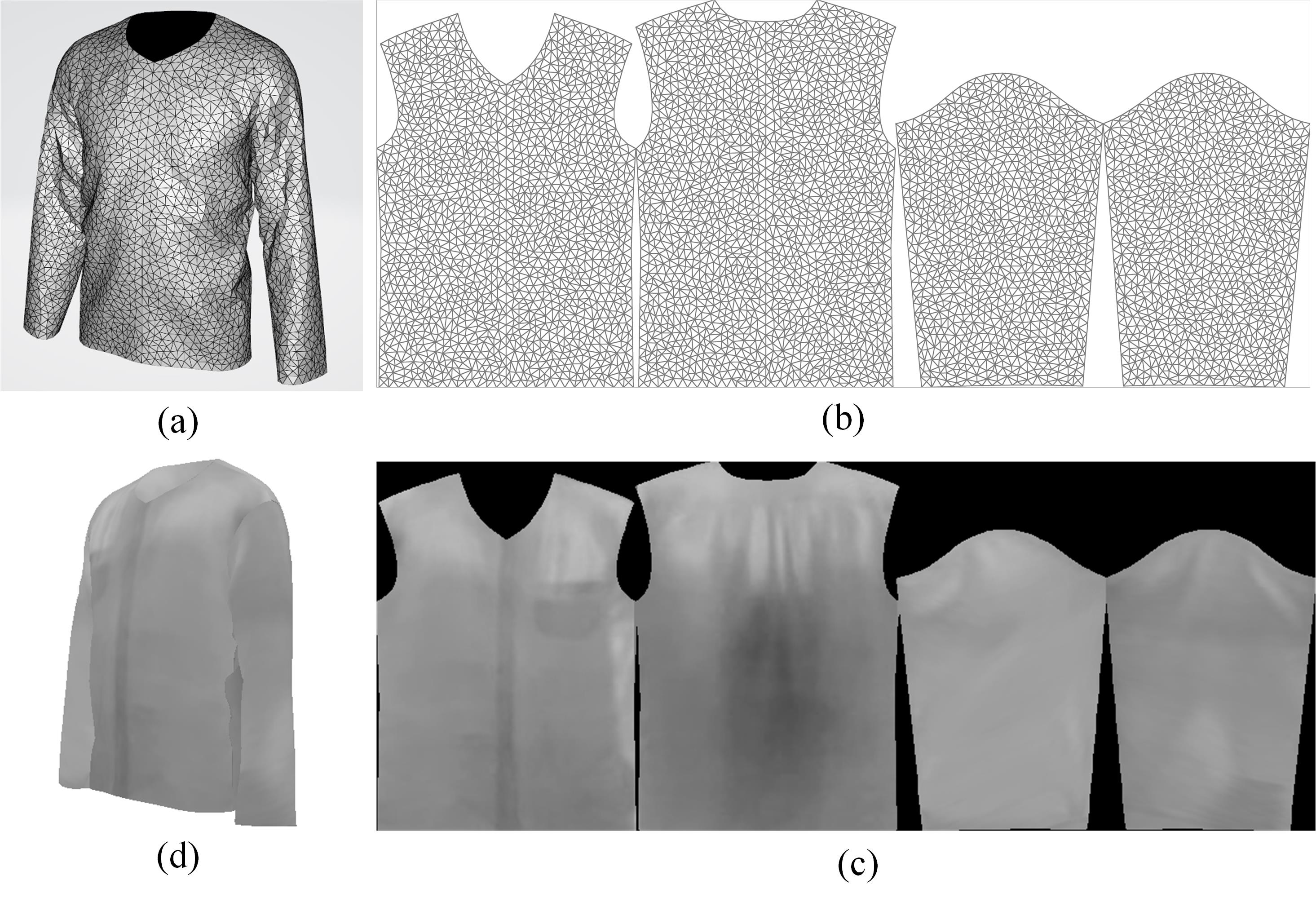} 
\caption{Construction of 3D clothing model with real infrared characteristics. (a) 3D clothing mesh model. (b) Faces map of 3D clothing model. (c) infrared texture map of 3D clothing model. (d) Rendered 3D infrared clothing model.}
\label{render}
\end{figure}

To address the above problem, we propose a method to use real infrared images to build the 3D human model 
and 3D clothing model. For example, Figure \ref{render}(a) shows a 3D mesh model of the clothing. Then we 
use infrared photos captured by infrared cameras to create ``skins'' for these mesh models. One challenge 
is how to paste these 2D photos to the surface of 3D mesh models. To address the above challenge, we 
first unfold the faces of 3D mesh models onto a 2D plane, which is called faces maps, as shown in 
Figure \ref{render}(b). After that, we divide these faces into different regions, such as arms, back, etc. by 
using the MAYA software. This process facilitates the alignment of 2D real infrared images with the 
3D mesh models.

Based on the faces map, we crop the infrared images to different parts and attach the cropped images 
onto the faces map using Photoshop software, and obtain the infrared texture map of 3D clothing model, 
as shown in Figure \ref{render}(c).  
We remark that the cropped infrared images may not perfectly align with the corresponding parts in the faces map. 
To address this issue, we handle it in two scenarios. First, if the captured image contains all the textures corresponding 
to the face map but exhibits slight shape mismatches, we use the distortion function in Photoshop to slightly deform the 
cropped images, ensuring better alignment with the corresponding face map areas. Second, if the captured image only contains 
partial textures corresponding to the face map (e.g., due to occlusion during capture), we first stitch images captured from 
different perspectives using Photoshop to create a complete texture. Then, we apply Photoshop's distortion function to deform the stitched image, ensuring it aligns with the face map.

Finally, we use the render of Pytorch3D \cite{ravi2020pytorch3d} to map the infrared texture map to the surface of the 3D clothing model. 
This method establishes a correspondence between the 3D mesh models and real 
infrared images. Figure \ref{render}(d) shows the rendered 3D clothes model based on real infrared characteristics.

\begin{figure}[tbp]
\centering
\includegraphics[width=1\columnwidth]{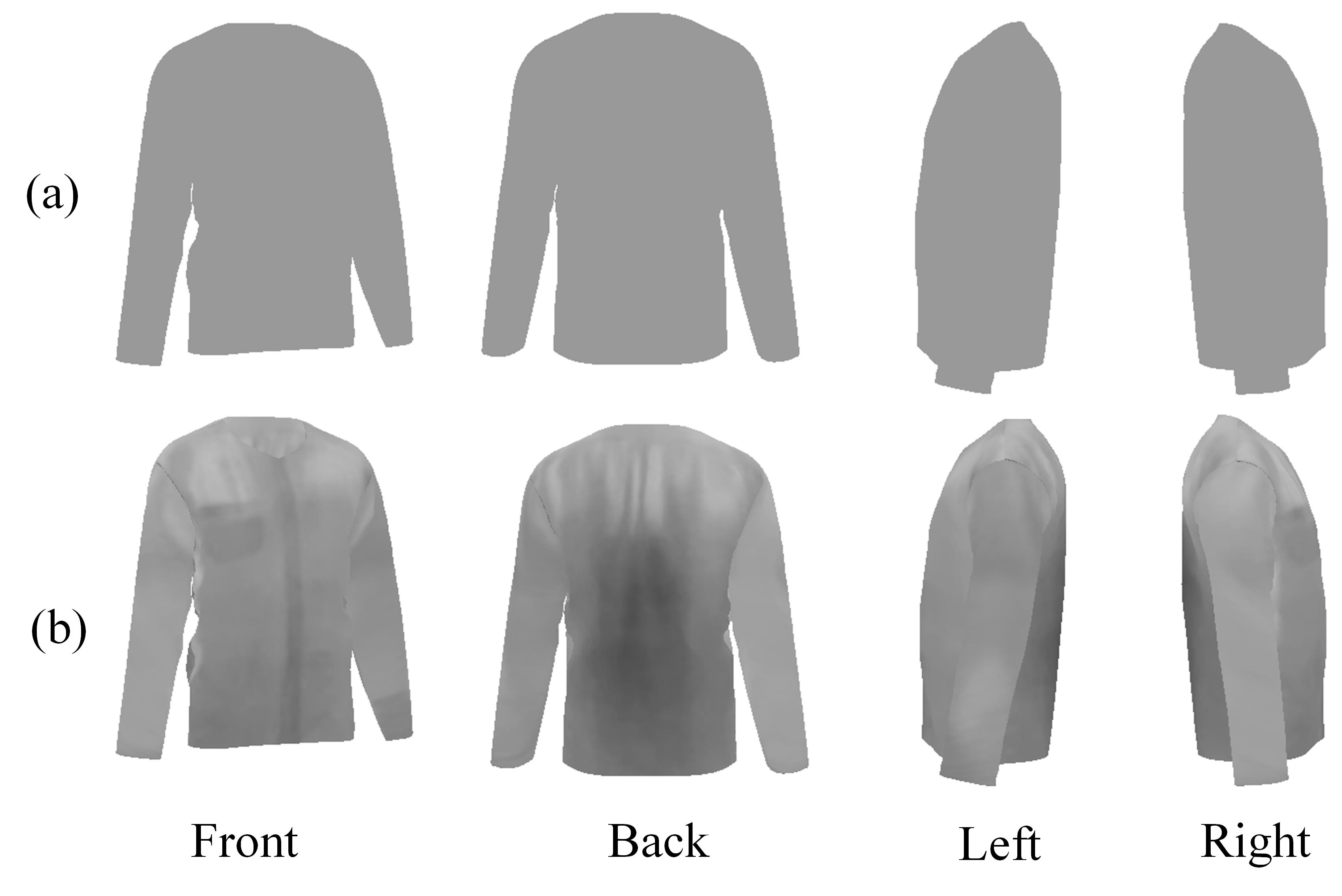} 
\caption{Comparison of 3D clothing models using the  (a) naive grayscale infrared modeling method and (b) real infrared characteristic modeling method (Ours). }
\label{model_compare}
\end{figure}

Since the 3D models we build are based on real-world infrared images, it minimizes the domain gap between 
the rendered 3D models and the real infrared images. This method helps to simulate physical adversarial attacks 
in the digital world more realistically than the naive grayscale infrared modeling method. Figure \ref{model_compare} shows the 
differences of the two kinds of infrared models. It indicates that the 3D infrared clothing model using our real 
infrared characteristic-based method has more infrared texture information and details than the grayscale infrared model.

For the generalization of the proposed method, our method can be applied to any 3D object because 
we only need to take infrared photos of these objects and paste them on the surface of 3D objects. 
Therefore, it has good generalization properties.

\subsection{3D infrared human body model}

2D images are uniformly represented by pixel arrays, while 3D models can be represented in multiple ways, such as meshes, point clouds, and voxel grids. We use meshes to build the proposed 3D model. Meshes are
collections of vertices and faces. Compared with point cloud and voxel representation, the 3D model represented by mesh typically has a smoother surface and richer
details. The human 3D mesh we constructed is shown in Figure \ref{main_process}, which consists of 257166 points and 128583 edges. Next, we build a ``skin" of the 3D model to get closer to the real-world person. The ``skin" is a figure
shown in Figure \ref{main_process}, which can be mapped to the surface of a 3D object using the render of Pytorch3D \cite{ravi2020pytorch3d}. 
The ``skin'' is created using the method described in Section \ref{sec:infared_model}. The render inputs scene information
(lights, camera, materials, geometry, and textures) and outputs an image. By changing the position of the camera, we can obtain rendered images of 3D models at multiple angles in different scenes. A rendered image of
the infrared human body model is shown in Figure \ref{main_process}.

\subsection{3D infrared adversarial clothing model} \label{sec:cloth_model}
In addition to the 3D human model, we also build a 3D clothing model to simulate people wearing clothes more realistically. The mesh of the clothing (Figure \ref{main_process}) is composed of 28092 points and 14064 edges.
The layout of the clothing is the proposed optimization target. We want to find a specific layout pattern that has an adversarial effect when it is rendered to the surface of the clothing. To facilitate physical implementation,
we design a pattern consisting of multiple black patches, as shown in Figure \ref{main_process}. In infrared images, the lower pixel value represents the lower temperature; thus, the black area represents the thermal insulation area,
and the other areas are without thermal insulation. We introduce the physical implementation of physical adversarial clothing in Section \ref{sec:physical implementation}. The combinations of multiple black patches form different patterns
on the surface of 3D clothes. We can change the pattern on the surface of 3D clothes by changing the position and orientation of the black patches. A rendered image of 3D clothing with an optimized pattern is shown in
Figure \ref{main_process} (upper panel).

\subsection{Simulating changeable infrared clothing characteristics in different times and places}\label{sec:change}
\begin{figure}[tbp]
\centering
\includegraphics[width=1\columnwidth]{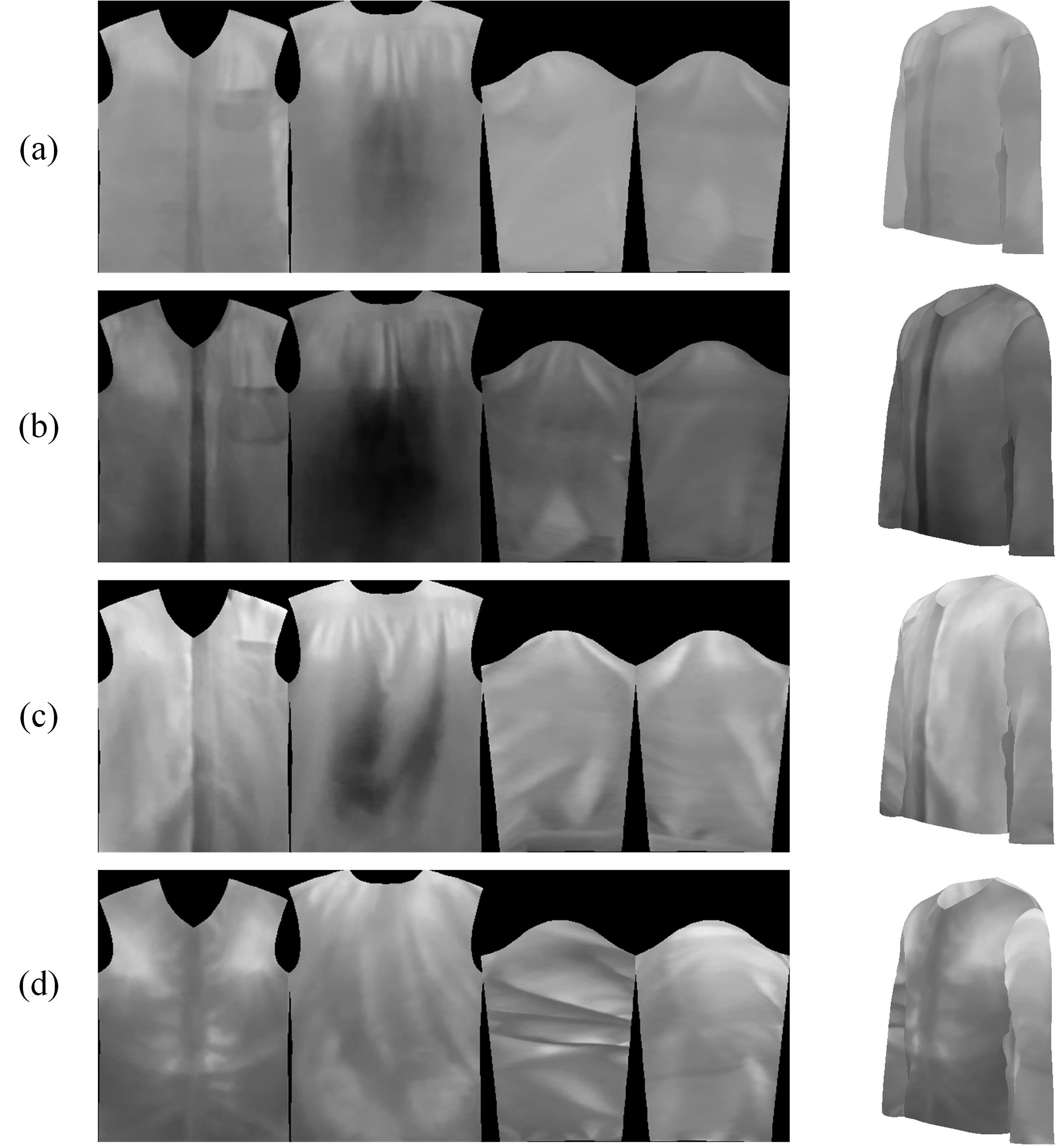} 
\caption{Different infrared characteristics of clothes at (a) day indoors, (b) day outdoors, (c) night indoors, and (d) night outdoors.}
\label{change}
\end{figure}

We found that the infrared characteristics of clothes also changed with different times and places. 
To simulate these different characteristics, we took infrared photos of the same clothes at different 
times (day and night) and different places (indoors and outdoors). After that, we used our proposed 
infrared 3D model building method (Section \ref{sec:infared_model}) to build infrared clothing models in different 
situations, as shown in Figure \ref{change}. During the optimization and testing process, we used the above 
method to randomly change the infrared texture map of 3D clothes to simulate the infrared 
characteristics of clothes at different times and places.

\subsection{Optimization of adversarial pattern in 3D infrared adversarial clothing} \label{sec:3D model method}

Figure \ref{main_process} shows the optimization process of the adversarial pattern in 3D infrared adversarial clothing. We assume that there are $N$ black squares with the same size on the pattern ${{P}_{\rm tex}}$. The variable $z$ records the
coordinates $\left( {{x}_{i}},{{y}_{i}} \right)$,$i=1,...,N$ and orientations ${{\alpha }_{i}},i=1,...,N$ of the center points of the $N$ patches. For ease of calculation, ${{x}_{i}}$ and ${{y}_{i}}$ are normalized
to $\left( 0,1 \right)$ according to the size of the pattern. The variable $z$ can be represented by the following equation:
\begin{equation}
    z=\left[ \begin{matrix}
        {{x}_{1}} & {{y}_{1}} & {{\alpha }_{1}}  \\
        {{x}_{2}} & {{y}_{2}} & {{\alpha }_{2}}  \\
        {} & ... & {}  \\
        {{x}_{N}} & {{y}_{N}} & {{\alpha }_{N}}  \\
     \end{matrix} \right].
\end{equation}
The pattern ${{P}_{\rm tex}}$ is generated by the variable $z$, and the generating function is $\it G$:
\begin{equation}
    {{{P}_{\rm tex}}}=\mathit{G}\left( z \right).
\end{equation}

As mentioned in Section \ref{sec:change}, we build four different texture maps for 3D infrared clothes in 
different times and places. During the optimization and testing process, we randomly change the infrared 
texture map of 3D clothes to simulate the infrared characteristics of clothes at different times and 
places. Let $\mathrm{Change}$ denote this process, so:
\begin{equation}
  {{P}_{\rm tex-C}}={\mathrm{Change}}\left( {{P}_{\rm tex}} \right).
\end{equation}

The goal of this study is to perform physical attacks; thus, we need to consider the perturbations in the real 
world. For example, when a person is moving, the positions of the black patches on the clothes may be 
shifted slightly. Therefore, we added some random noise to the coordinates and orientations of these 
patches. In addition, when the ambient temperature changes, the pixel value of the black patch in the 
infrared image may change. Therefore, we set the pixel value of the black patch to vary randomly 
within a certain range. To simulate the diverse textures of real clothing, we introduce random 
grayscale variations on the clothing surface during training. The method to simulate 
disturbance in the physical world is called expectation over transformation 
(EOT) \cite{conf/icml/AthalyeEIK18}. The pattern after EOT transformation is ${{P}_{\rm tex-CE}}$:
\begin{equation}
  {{P}_{\rm tex-CE}}={\rm EOT}\left( {{P}_{\rm tex-C}} \right).
\end{equation}

\begin{figure*}[htbp]
\centering
\includegraphics[scale = 0.6]{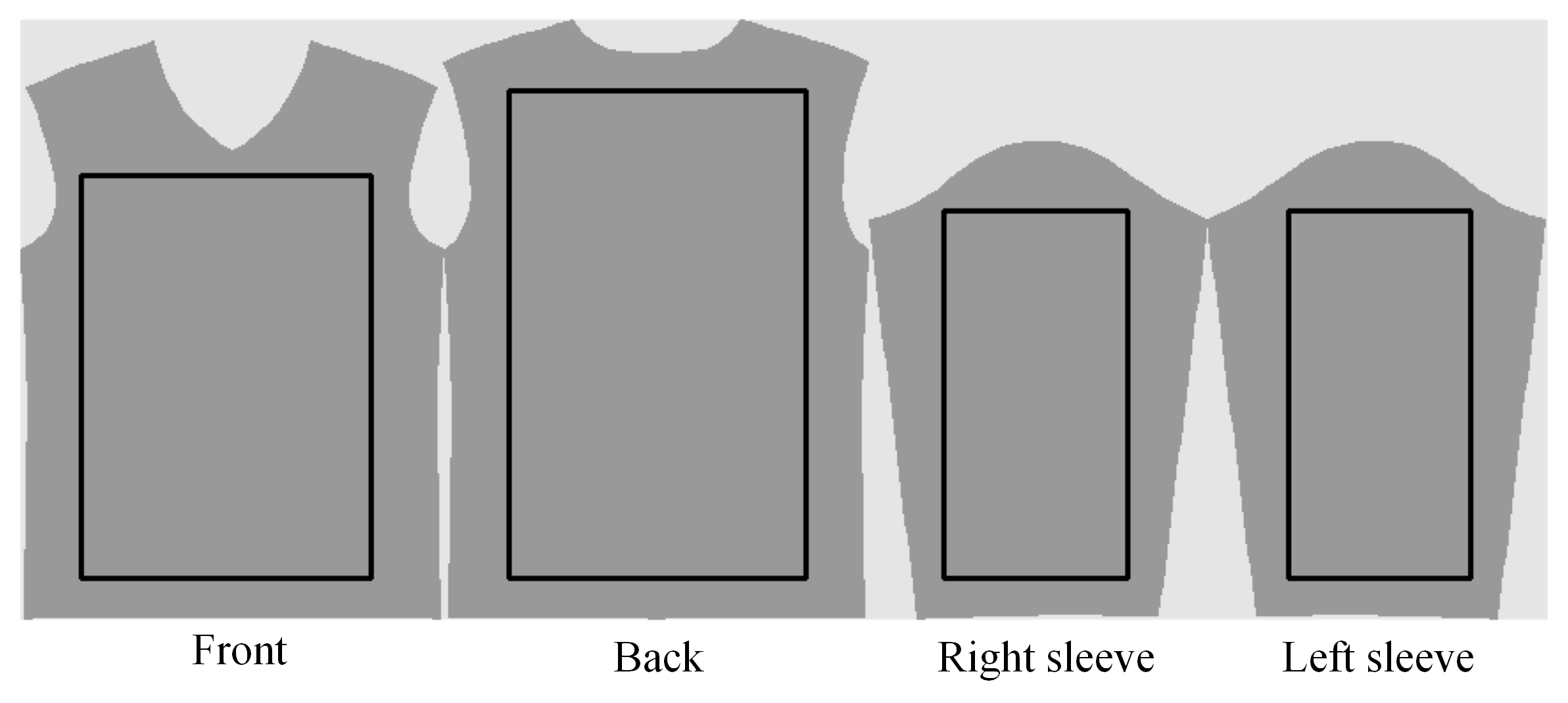} 
\caption{Optimization area mapped to the surface of the 3D clothing model. The dark gray area is the mapping area. The center coordinates of the
patches are constrained within the black bounding box areas.}
\label{opt_area}
\end{figure*} 

The render in Pytorch 3D maps a 2D pattern to the surface of a 3D clothing mesh and generates a rendered image. Figure \ref{opt_area} shows the mapping relationship between the area on the pattern and the 3D clothing
surface in our experiment. It looks like we have cut the 3D clothes into four fabrics and put them on a 2D plane. 
Specific areas in the pattern (marked by dark gray in Figure \ref{opt_area}) will be mapped to the surface of the 3D clothing. The center coordinates of
the patches were constrained within the black bounding box area, which ensures that black patches are in the valid mapping area.
The scene parameters $\varphi $ of the renderer $R$ include lights, camera and
so on. With the input of 3D clothing mesh ${{M}_{\rm cloth}}$, pattern ${{P}_{\rm tex-CE}}$, and scene parameters $\varphi $, the render $R$ outputs the rendered image ${{I}_{\rm cloth}}$:
\begin{equation}
    {{I}_{\rm cloth}}=\mathit{R}\left( {{M}_{\rm cloth}},{{P}_{\rm tex-CE}},\varphi  \right).
\end{equation}

We combine the 3D person model and 3D clothing model, in other words, we let the 3D person ``wear'' the 3D clothing. The 3D person model includes 3D person mesh ${{M}_{person}}$ and skin image ${{P}_{skin}}$. We use
the render to generate the rendered image of a man wearing the clothing:
\begin{equation}
    {{I}_{\rm man}}=\mathit{R}\left( {{M}_{\rm person}},{{P}_{\rm skin}},{{I}_{\rm cloth}}  \right).
\end{equation}

To simulate the attack effect in different scenes, we paste the rendered images ${{I}_{\rm man}}$ to different background images ${{I}_{\rm back}}$ and obtain the pasted image ${{I}_{\rm paste}}$:
\begin{equation}
    {{I}_{\rm paste}}=\mathrm{Paste}\left( {{I}_{\rm man}},{{I}_{\rm back}} \right).
\end{equation}

We input the pasted image ${{I}_{\rm paste}}$ into the object detection model $f$ parameterized by $\theta $. The output of the detection model includes the bounding
box ${{f}_{\rm pos}}\left( {{I}_{\rm paste}},\theta  \right)$, the object score ${{f}_{\rm obj}}\left( {{I}_{\rm paste}},\theta  \right)$, the class score ${{f}_{\rm cls}}\left( {{I}_{\rm paste}},\theta  \right)$.
The goal is to make the detector unable to detect the person. Then, we try to lower ${{f}_{\rm obj}}\left( {{I}_{\rm paste}},\theta  \right)$ as much as possible.

Considering that there are some other persons in the background images ${{I}_{\rm back}}$, the detector may output multiple predicted bounding boxes $B=\left\{ {{b}_{j}} \right\}_{j=1}^{T}$. To avoid the interference of
these background persons, we only keep the bounding box closest to the ground truth of the 3D model (i.e., the box that has the max object score when its intersection over union (IOU) with the ground truth (GT) is above
a certain threshold ${\varepsilon }_{\mathrm{IOU}}$). We define the following loss function ${{L}_{\det }}$:
\begin{equation}
    {{L}_{\rm det }}=\max_{j\in\mathcal{J} }{f_{\mathrm{obj}}^{(j)}\left( {I_{\mathrm{paste}}},\theta  \right) }, 
\end{equation}
where:
\begin{equation}
    \mathcal{J} = \left\{1,2,...,T\right\}\cap\left\{j\vert\mathrm{IOU}\left(\mathrm{gt},{{b}_{j}} \right)>{{\varepsilon }_{\mathrm{IOU}}}\right\}.
\end{equation}

The optimization target is the positions and orientations of black patches on the surface of 3D clothes. We want to optimize a pattern with a special arrangement of black square patches on the clothes that can hide from detectors.
These black patches correspond to physical insulation materials. To facilitate the physical implementation, we try
to minimize the overlap between different patches. We propose a new loss function, $L_{\rm dist}$, which aims to increase the average distance between every two patches as much as possible within a certain
range. $\left( {{x}_{i}},{{y}_{i}} \right)$ and $\left( {{x}_{j}},{{y}_{j} } \right)$ are the coordinates of the center point of every two patches, and $-d$ is the lower bound of the $L_{\rm dist}$
to prevent the distance between every two patches from being too large:
\begin{equation}
    {{L}_{\rm dist}}=\max \left( -\frac{\sum\limits_{\forall i\ne j}^{N}{\sqrt{{{\left( {{x}_{i}}-{{x}_{j}} \right)}^{2}}+{{\left( {{y}_{i}}-{{y}_{j}} \right)}^{2}}}}}{N\times \left( N-1 \right)/2},-d \right).
\end{equation}

The loss function of the detector is:
\begin{equation} \label{loss_v3}
    L={{L}_{\det }}+\lambda {{L}_{\rm dist}},
\end{equation}
where $\lambda $ is a hyperparameter that is determined empirically.

For ensemble attacks, we want to lower the object scores of multiple detectors concurrently. We assume that there are $K$ detectors, and the object score output by each detector is
$L_{\det }^{\left( k \right)}, k=1,..,K$. The loss function of the ensemble attack is:
\begin{equation}
    {{L}_{\rm ensem}}=\sum\limits_{k=1}^{K}{L_{\rm det }^{\left( k \right)}}+\lambda {{L}_{\rm dist}}.
\end{equation}
We use the backpropagation algorithm to update the variable $z$ according to the loss function and update the pattern of the clothes ${{P}_{\rm tex}}$.

\subsection{Physical implementation}\label{sec:physical implementation}
Thermal insulation materials block the thermal radiation of the human body; thus, the thermal-insulated areas appear black in the infrared images. We compared the thermal insulation performance of two common fabrics
(cotton and polyester), a common tape (rubber tape), two thermal insulation tapes (polyimide, Teflon), and a new material - aerogel. We found that the aerogel has good
thermal insulation properties that are stable over time. We purchased a piece of $1.5m\times1.5m\times6mm$ aerogel cloth from ZhongPu Company (Hebei, China) and cut it into many $6cm\times6cm\times6mm$ aerogel patches.

We can choose a piece of physical clothing that is similar to the shape and size of the 3D clothes created in the digital world.
According to Figure \ref{opt_area}, we draw four rectangles on the clothes.
In each rectangle, we set up a Cartesian coordinate system whose origin is at the center of the box, and the X and Y axes are parallel to the sides of the rectangular box. 
This process is performed on both the digital and physical clothes. Then, the positions and orientations of patches in the digital world are mapped to the physical world.
This mapping does not need to be very precise because the optimization of the patch layout pattern in the digital world has considered random perturbations (see EOT transformations in Section 3.3).
The aerogel patches are fixed at those positions. 
The areas outside the rectangular boxes on the clothes are not our optimization areas; thus, we allow some differences between the physical clothes and the 3D model in these outside areas.

In our experiment, we used a sunscreen coat that had a zipper on the front (Figure \ref{main_process}, bottom), which was helpful for placing the
aerogel patches inside the coat. The collar of the coat was different from the collar of the 3D model clothes, but as
explained above, this difference is allowed. We used tape and thread to fix the aerogel patches on the coat.

Infrared imaging has the penetrating ability. Even if these aerogel patches are on the inner side of clothes, the pattern on the clothes can
be displayed under infrared imaging. The physical adversarial clothes look like ordinary clothes in a visible light view.

\section{Experiments}
\subsection{Dataset}
We chose the FLIR\_ADAS\_1\_3 \cite{FLIR_ADAS} dataset released by FLIR Corporation. FLIR\_ADAS\_1\_3 provides annotated infrared images for training and testing object detection tasks. The dataset contains a total of 10228 infrared
images. We used 7160 images as the training set and 3068 images as the test set. These images were taken on the highways and streets of Santa Barbara, USA, from November to May. The infrared camera was a FLIR
Tau2 (NETD$<$60 mK, FPA $640\times512$, $13 mm f/1.0$). Infrared images were manually annotated and contain four types of objects: people, dogs, cars, and bicycles. 
We used these real-world infrared images to finetune the infrared object detector and as the background for the
3D models.

\subsection{Target detectors}
We chose the latest YOLOv9 \cite{wang2024yolov9} detector as our primary target detector, and finetuned the officially 
pre-trained YOLOv9 detector on the FLIR\_ADAS\_1\_3 dataset. The average precision for person category of 
the finetuned detector was 0.98 on the training set and 0.95 on the test set. After attacking YOLOv9 in the 
white-box setting, we evaluated the transferability of our method to other unseen detectors such as 
DETR\cite{carion2020end}, DINO\cite{zhang2022dino}, DDQ\cite{zhang2023dense}, DAB-DETR\cite{liu2022dabdetr}, 
YOLOv7\cite{wang2023yolov7}, YOLOv8\cite{sohan2024review}, YOLOv11\cite{khanam2024yolov11}, Mask-RCNN\cite{he2017mask}, 
and Cascade RCNN\cite{cai2019cascade} in the black-box setting. 
These models are provided by the mmdetection library \cite{mmdetection}. 

\subsection{Evaluation Metrics}
We used the attack success rate (ASR) as the evaluation metric, which is commonly used in many 
researches \cite{journals/corr/abs-1910-11099,conf/cvpr/EykholtEF0RXPKS18,hu2022cvpr,zhu2021fooling,zhu2022cvpr,conf/ijcai/XiaoLZHLS18} 
in this area. In digital attacks, we defined the annotation of the 3D person model 
(digital attack) or real persons (physical attack) as ground truth (GT), and used the intersection over 
union (IOU) method to calculate the detection accuracy. Because the background image (FLIR\_ADAS dataset 
or real photos) contains other persons, we defined the evaluation metric ASR as the number of 
undetected 3D human models divided by annotated 3D human models (digital attack) or undetected persons divided by annotated 
persons (physical attack). When the detector outputs multiple bounding boxes, we selected the box with the 
IOU with GT greater than 0.5 and the highest object score with the threshold 0.7 as the detection result. 

\subsection{Simulation of physical attacks}
We conducted all experiments on 8 2080TI GPUs. 

\subsubsection{Attack YOLOv9 in the digital world} \label{sec:digital attacks}


\begin{figure}[htbp]
\centering
\includegraphics[width=1\columnwidth]{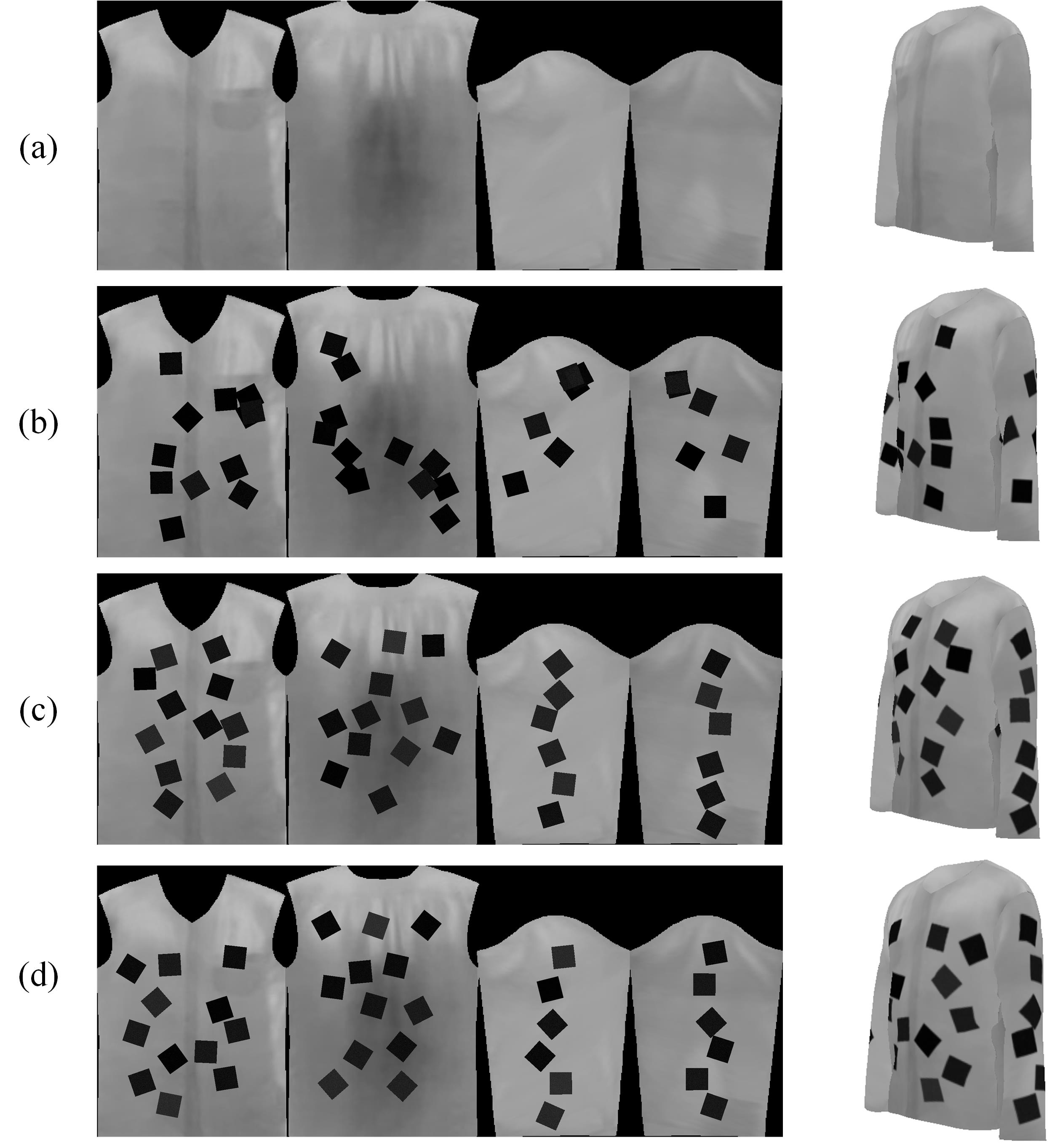} 
\caption{Different patterns of the texture map and rendered 3D clothing model. 
(a) Clean pattern. (b) Random position pattern. (c) Adversarial pattern optimized by YOLOv9. 
(d) Adversarial pattern optimized by ensemble model.}
\label{v3_texture}
\end{figure}

\begin{figure*}[htbp]
\centering
\includegraphics[scale=0.45]{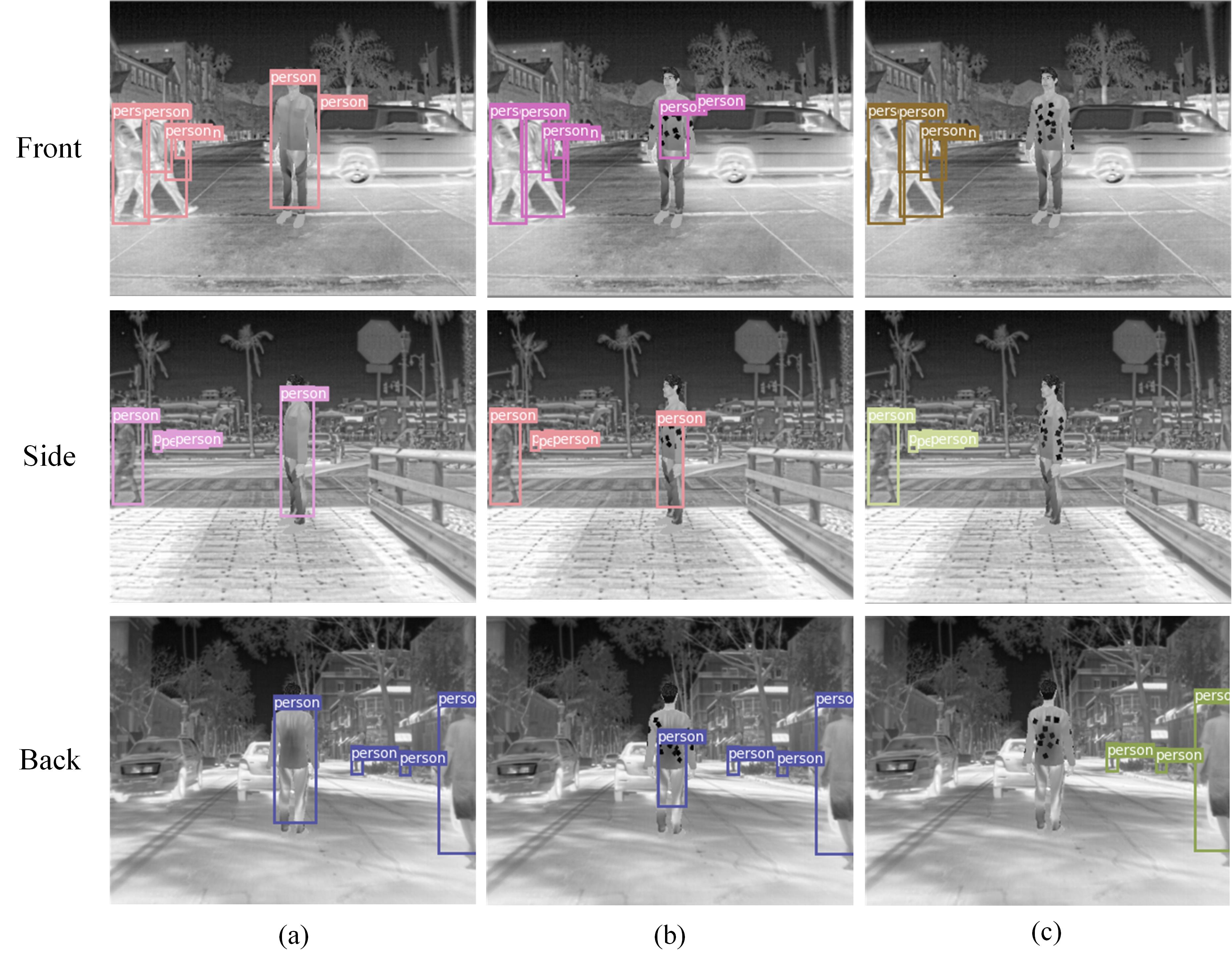} 
\caption{Examples of digital attacks. (a) Ordinary clothing. (b) Random pattern. (c) Optimized pattern. }
\label{digital_examples}
\end{figure*} 

\begin{table}[htbp] 
\caption{Evaluation of digital attacks}\label{tab_digital_attacks}
\centering
\begin{tabular}{cc}
\toprule[1.1pt]  
Pattern of 3D clothing model & ASR\\
\midrule  
Ordinary clothing &  4.65\%  \\
Random pattern & 38.19\% \\
Optimized pattern & 86.38\% \\
\bottomrule[1.1pt] 
\end{tabular}

\end{table}

As mentioned in Section \ref{sec:3D model method}, the pattern ${{P}_{\rm tex}}$ of 3D clothing contains $N$ black patches. We took $N=36$ in our experiments (see Section \ref{sec:patch num} for the study of other values). According to the ratio of different
areas on the clothes, we mapped 12 (N/3) patches onto the front of the clothes, 12 (N/3) patches onto the back of the clothes, and 6 (N/6) patches onto the left and right sides of the clothes. The center coordinates of the
patches were constrained within the bounding box (the area marked by the black bounding box in Figure \ref{opt_area}) during optimization and were randomly initialized. This ensured that black patches were always in the valid mapping area (marked by dark gray area
in Figure \ref{opt_area}). The EOT transformations include random noise ([-0.01,0.01]), random shifting ([-0.5 cm, 0.5 cm]),
random rotation ([-0.03 radians, 0.03 radians]), random brightness ([0, 0.1]), and random grayscale variations ([-0.3, 0.3]).

We used the differentiable render in Pytorch3D to map the pattern on the 3D clothing surface. Then we combined 
the 3D clothing model and 3D person model, that is, we let the 3D person ``wear” the 3D clothing. We set the height of the 3D human model to $1.75m$ and the size of the black patch to be $6cm\times6cm$ 
(see Section \ref{sec:patch size} for the study of other sizes). In the digital world, the distance between the camera and the human body varied from $2.5m$ to $5m$, and the angle varied randomly between 0-360 degrees. 
We obtained rendered images at different angles and distances. We randomly selected the background images from FLIR\_ADAS\_1\_3 training set, and pasted the rendered images to the center of the background images, 
so that we could switch various scenes for attack simulation. 

We input the images with the background into the YOLOv9 detector. We used the Adam optimizer with momentum. The optimizer used the backpropagation algorithm
to update the variable $z$, which records the positions and orientations of the 36 black patches in the pattern, and then updates the patch layout pattern. The
hyperparameter $\lambda$ in Equation (\ref{loss_v3}) was 0.1 (the sensitivity of $\lambda$ is analyzed in Section \ref{sec:lambda}). The initial learning rate was 0.03.  If the loss function did not drop after 50 iterations, 
then the learning rate gradually decreased until 0.001, and then the optimization process ended at that time.  We obtained a stable pattern using this method. 
It took approximately 6 hours to obtain the optimization result on our GPU server for each experiment. Figure \ref{v3_texture}(c) shows the optimized pattern.

Next, we rendered the optimized pattern on the surface of the 3D clothing model, and tested its attack effect on the test set of FLIR\_ADAS\_1\_3. The settings of the testing process were the same as the optimization 
process. For fair comparison, we used the ordinary clothing pattern (Figure \ref{v3_texture}(a)) and random postion pattern (Figure \ref{v3_texture}(b), with 36 randomly placed black patches) for control experiments. These different patterns were rendered on the same 3D model, and the 
rendered images are input to the YOLOv9 detector. We defined the annotation of the 3D person model as ground truth (GT), and used the intersection over union (IOU) method to calculate the detection accuracy. 
ASR is used as the attack evluation metric.

Table \ref{tab_digital_attacks} shows the results. 
The ASR of the optimized pattern (86.38\%) against YOLOv9 was obviously better than that of
random patten (38.19\%) and clean pattern (4.65\%).
Figure \ref{digital_examples} shows three types of clothes at three different viewing angles.

\subsubsection{Effect of the number of black patches} \label{sec:patch num}
As described in Section \ref{sec:digital attacks}, the pattern included 36 black patches. Then, we 
tried different patterns, which included 6, 12, 24, 36, 48, and 60 black patches, to investigate 
how the number of black patches in the clothing pattern
affects the attack performance of 3D clothing. The optimization and testing process were the same 
as described in section \ref{sec:digital attacks}. The results are shown in Table \ref{tab_patch_number}. 
The results indicated that when the number of patches increased
from 6 to 60, the attack performance increased first and then decreased. 
When $N$ was 36, the attack effect was the best. The decreasing attack effect with many more patches 
is consistent with our recent study \cite{zhu2022cvpr} that 
fully insulated clothing had only weak attack effect.

\begin{table*}[htbp] 

\caption{Effect of the number of black patches }\label{tab_patch_number}
\centering
\begin{tabular}{c|cccccc}
\toprule[1.1pt]  
The number  & 6  & 12   & 24   & 36  & 48   & 60  \\
\midrule  
ASR &  22.38\% & 50.32\% & 72.18\% & 86.38\% & 85.25\% & 83.73\% \\
\bottomrule[1.1pt] 
\end{tabular}

\end{table*}

\begin{table*}[htbp] 
\centering
\caption{Effect of black patch size}\label{tab_patch_size}
\begin{tabular}{c|cccccc}
\toprule[1.1pt]  
Black patch size (cm)  & 2  & 4   & 6   & 8   & 10   & 12  \\
\midrule 
ASR &  51.30\% & 70.07\% & 86.38\% & 85.95\% & 81.31\% & 78.12\% \\
\bottomrule[1.1pt] 
\end{tabular}

\end{table*}

\begin{figure*}[tbp]
\centering
\includegraphics[scale = 0.5]{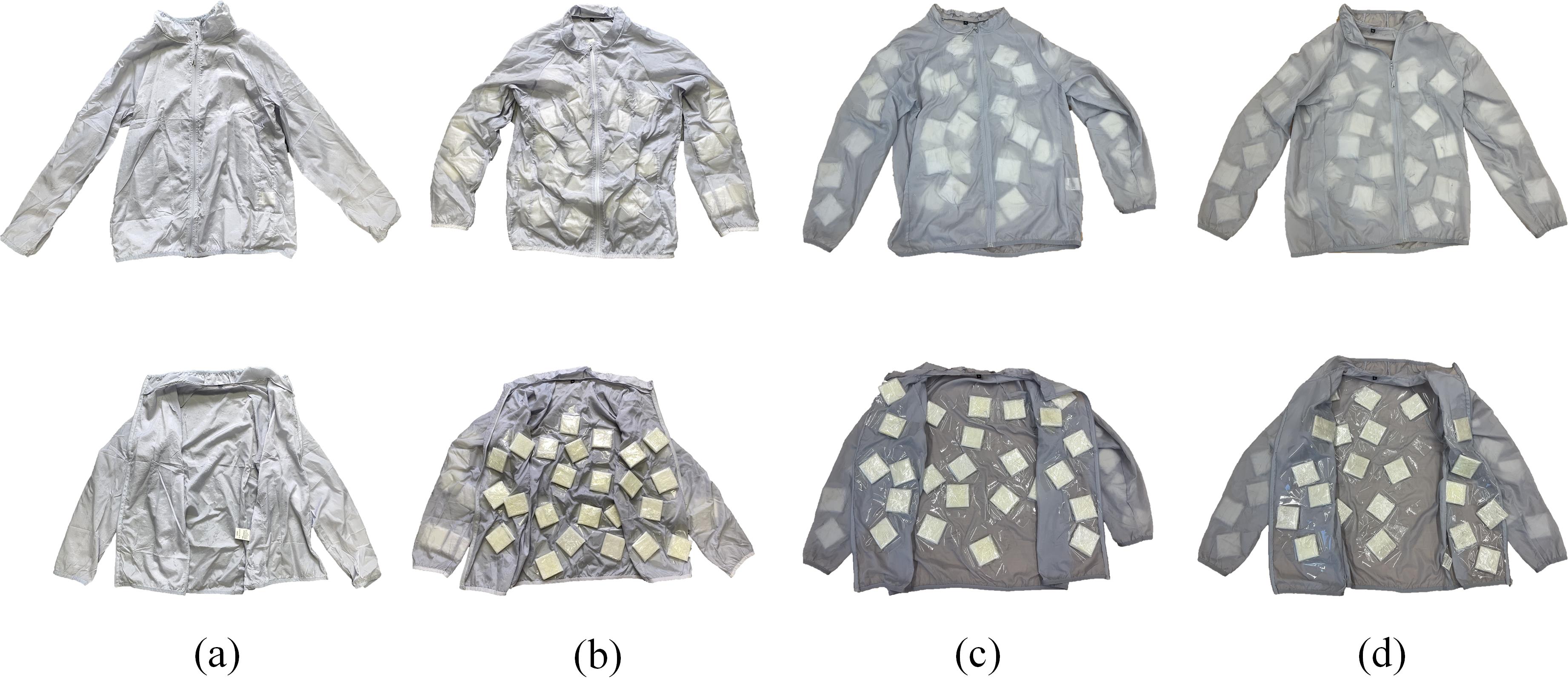} 
\caption{Physical adversarial clothing. (a) Ordinary clothing. (b) Random pattern clothing. (c) Adversarial clothing optimized with respect to YOLOv9. (d) Adversarial clothing optimized with respect to the ensemble model.} 
\label{phy_cloth}
\end{figure*} 

\begin{table*}[htbp] 
\centering
\caption{Sensitivity analysis of $\lambda$ in the loss function}\label{tab_lambda}
\begin{tabular}{c|cccccc}
\toprule[1.1pt]  
The value of $\lambda$  & 0  & 0.01   & 0.05   & 0.1   & 0.5   & 1  \\
\midrule  
ASR &  72.66\% & 75.35\% & 83.19\% & 86.38\% & 80.10\% & 76.61\% \\
\bottomrule[1.1pt] 
\end{tabular}

\end{table*}

\subsubsection{Effect of black patch size} \label{sec:patch size}
In Section \ref{sec:digital attacks}, the size of the black patch was $6cm\times6cm$ in 3D modeling. Then, we studied the attack effect of patterns with patch sizes of 2 cm, 4 cm, 6 cm, 8 cm, 10 cm, and 12 cm. Other settings were the
same as in Section \ref{sec:digital attacks}. The results are shown in Table \ref{tab_patch_size}. 
The results indicated that with the increase in patch size, the ASR first increased and then decreased.
The size of $6cm \times 6cm$ was the best.

\subsubsection{Sensitivity analysis of $\lambda$ in the loss function} \label{sec:lambda}
The hyperparameter $\lambda$ in Equation (\ref{loss_v3}) balances $L_{det}$ and $L_{dist}$. 
We set $\lambda=0.1$ by default. To analyze the sensitivity of $\lambda$, we set $\lambda$ to 
be 0, 0.01, 0.05, 0.1, 0.5, and 1. We kept other 
optimization and testing settings same as described in Section \ref{sec:digital attacks}. 
Table \ref{tab_lambda} shows the results. 
The results indicated that when the $\lambda$ value 
increased from 0 to 0.1, the ASR increased. 
But when $\lambda$ was greater than 0.1, the ASR decreased.  
$\lambda = 0.1$ was the best.

\begin{figure}[htbp]
\centering
\includegraphics[width=0.8\columnwidth]{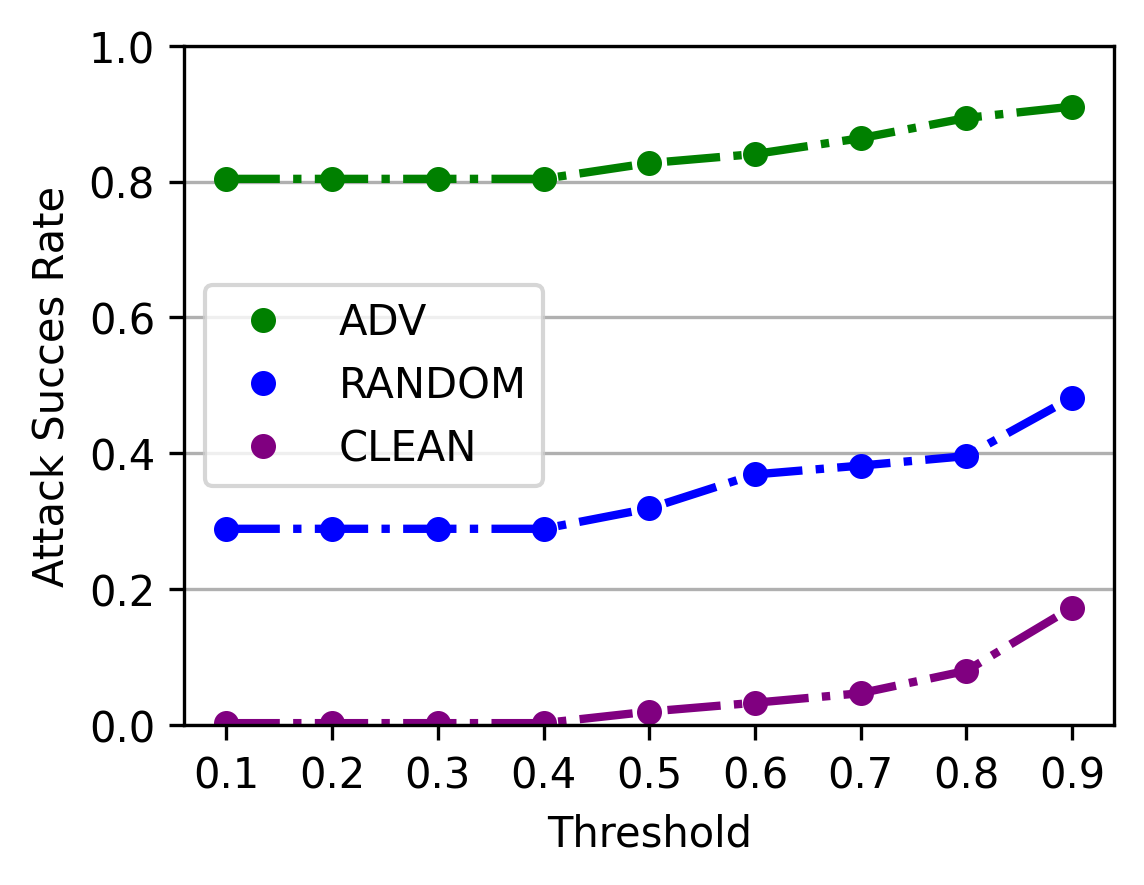} 
\caption{ASRs at different object score thresholds.}
\label{asr_thresh}
\end{figure}

\subsubsection{Effect of detection threshold} \label{sec:thresh}
We tested different object score thresholds and calculated the ASRs at different thresholds 
(from 0.1 to 0.9). We plot the curve of ASR with different thresholds, as shown in Figure \ref{asr_thresh}.

The results show that the ASR will increase as the detection threshold increases. 
Despite the change in threshold, the ASRs of our adversarial pattern are obviously higher than 
that of random pattern and clean pattern.  However, we found that 
if the threshold is too high, the recall of clean pedestrian detection will decrease. If the threshold 
is too low, the false detection rate will increase. Therefore, a typical threshold is 0.75 in the 
original YOLOv9 paper\cite{wang2024yolov9}, which is similar to 0.7. Besides, in pedestrian attacks, the commonly used 
settings are generally 0.6 or 0.7, which is the same in many previous works \cite{journals/corr/abs-1910-11099,hu2022cvpr,zhu2021fooling,zhu2022cvpr}. 

\subsubsection{Evaluation on different datasets}
We evaluated the proposed method on the other two infrared pedestrian datasets: KAIST\cite{choi2018kaist} 
and FLIR\_ADAS\_v2 \cite{FLIR_v2}. We used these datasets to finetune the object detector YOLOv9 and as the 
background for 3D model simulation. The settings of other experiments are consistent with 
Section \ref{sec:digital attacks}. The results are shown in Table \ref{tab_dataset}. Our method 
achieved good results on different datasets, and the difference in ASR is not significant. 
It indicates the generalizability of our method across different datasets.

\begin{table*}[htbp]
\centering
\caption{ASRs on different datasets.}\label{tab_dataset}
\begin{tabular}{cccc}
\toprule[1.1pt]  
\diagbox{Method}{Dataset} & FLIR\_ADAS\_1\_3 & KAIST & FLIR\_ADAS\_v2 \\
\midrule  
Clean & 4.65\% & 6.59\% & 5.31\%  \\
Random & 38.19\% & 36.71\% & 38.27\%  \\
Adversarial & 86.38\% & 83.27\% & 85.58\%  \\
\bottomrule[1.1pt] 
\end{tabular}

\end{table*}

\subsection{Evaluation of attacks in the physical world}
In our previous work \cite{zhu2022cvpr}, we tested the thermal insulation properties of five materials, and found that the aerogel material had better thermal insulation properties than other materials and remained stable over time.
We therefore used this material in this study.

\begin{figure*}[htbp]
\centering
\includegraphics[scale = 0.11]{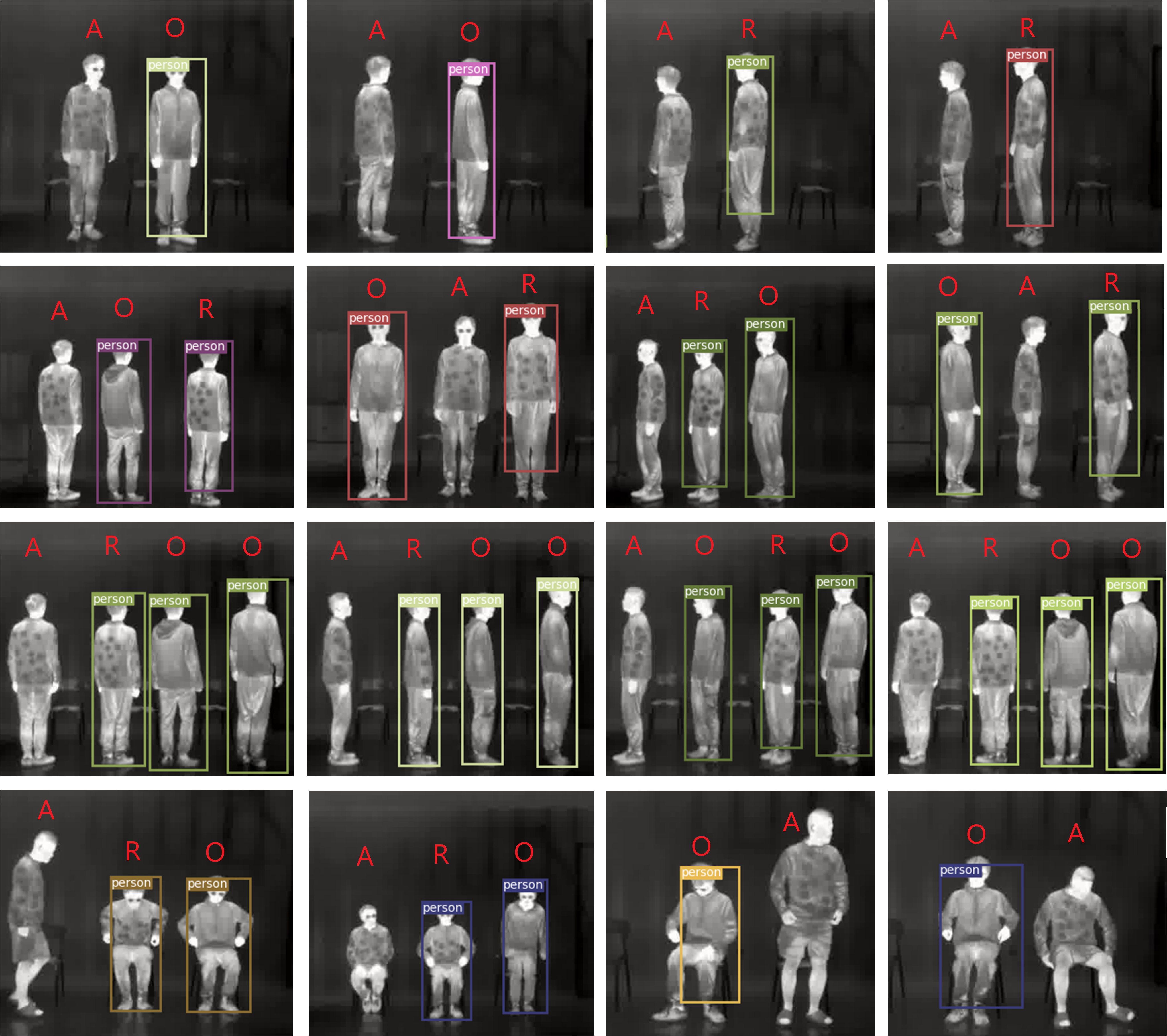} 
\caption{Examples of physical infrared attacks. A: adversarial clothing. R: random pattern clothing. O: ordinary clothing.} 
\label{phy_examples}
\end{figure*}

\subsubsection{Attack YOLOv9 in the physical world} \label{sec:physical attacks}
We manufactured physical adversarial clothing based on the optimized 3D clothing model, and the manufacturing process is described in Section \ref{sec:physical implementation}. For a fair comparison, we used random pattern clothing (the clothing with 36 randomly placed aerogel
patches) and ordinary clothing for the control experiments. All these clothes had the same size. Figure \ref{phy_cloth} shows these physical clothes.

\begin{table}[bp] 
\caption{Evaluation of physical attacks} \label{tab_phy_attacks} 
\centering
\begin{tabular}{ccc}
\toprule[1.1pt]  
Physical clothing & ASR (indoor) & ASR (outdoor) \\
\midrule  
Ordinary clothing &  7.12\% &  8.38\% \\
Random pattern clothing & 26.53\%  &  23.03\%\\
Adversarial clothing & 80.11\%  &  76.85\%\\
\bottomrule[1.1pt] 
\end{tabular}
\end{table}

We tested the attack effect of these clothes in the physical world. We invited 8 volunteers. The human-related experiments have been approved by the Ethics Committee of Tsinghua University. The volunteers wore the 
adversarial clothing, the random pattern clothing, or ordinary clothing. We used the infrared camera FLIR T630sc to capture these volunteers in multiple scenes indoors and outdoors. They could stand, sit or 
in any posture they want. The angle from the camera to human body varied between 0-360 degrees, and the distance was between 1-12 meters. We input the infrared photos into the YOLOv9 detector with a threshold 0.7. Some 
examples are shown in Figure \ref{phy_examples}. The results showed that people wearing adversarial clothing were not detected, while people wearing the random pattern clothing, or ordinary clothing were detected. 
This indicated the effectiveness of the proposed method. See Supplementary Video for the demo.

\begin{figure}[bp]
\centering
\includegraphics[width=1\columnwidth]{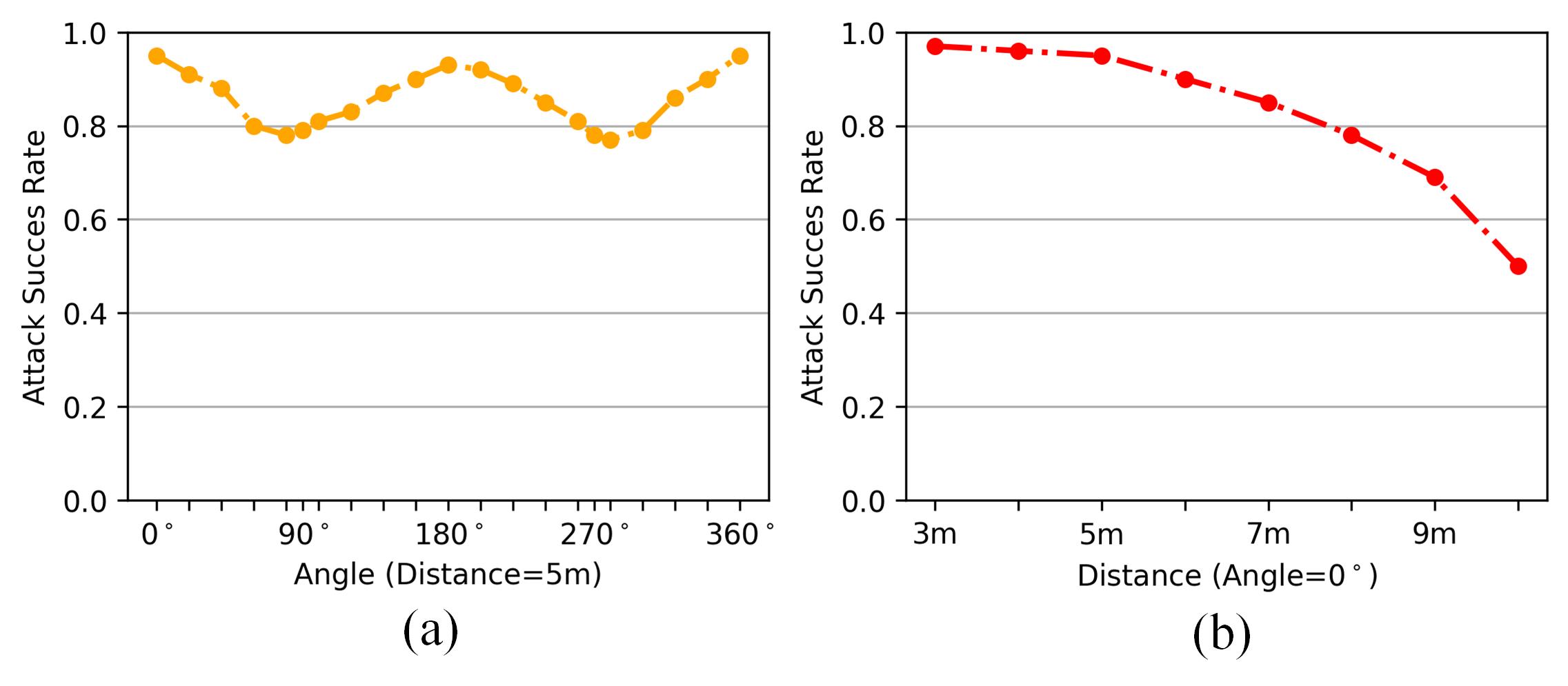} 
\caption{Analysis of physical attack effects at (a) different angles and (b) different distances. }
\label{angle_dist}
\end{figure}


To quantitatively evaluate the attack effect, we recorded 100 videos. Half of the videos were recorded indoors; the other half were recorded outdoors.  We fixed the position 
of the camera, and the distance between the volunteer and the camera was 3$m$, 5$m$, or 7$m$. When we recorded the videos, the volunteers rotated in situ at a constant speed. For a fair comparison, 
the same volunteers wore ordinary clothes, the random pattern clothing, and the adversarial clothing, respectively at the same position. We sampled these videos at 3 frames 
per second and obtained 6000 images. The images that include each kind of clothing took 1/3 of the total images. We manually annotated all images and input them into YOLOv9. Table \ref{tab_phy_attacks}
shows the results. 
The results indicated that, the ASRs of adversarial clothing against YOLOv9 were 80.11\% (indoor) and 
76.85\% (outdoor) in the physical world, while ASRs of the random pattern clothing and ordinary clothing 
were only 26.53\% and 7.12\%, respectively indoors, and 23.03\% and 8.38\%, respectively outdoors 
in the physical world.
The results showed the effectiveness of our physical attacks compared to the control experiments. 
We also noticed that physical attacks were more difficult in outdoor scenes, possibly because the outdoor environment is more complex. 
For example, when the weather is hot (e.g. higher than 30 degrees Celsius), the temperature of the aerogel patches is close to that of the human body. Therefore, the pixel values of black patches are close to those of other areas in infrared images of adversarial clothing, 
and the adversarial pattern is not shown clearly.

We then analyzed the attack performance of adversarial clothing from different angles. The distance 
between the volunteers and the camera was fixed at 5 meters. The volunteers rotated in situ at a 
constant speed. We took 11 
sample points between 0-360 degrees and marked key angles on the ground. We selected 100 images at 
each angle for statistics, and used the ASR as the evaluation method. We input these images into 
the YOLOv9 detector 
with the threshold 0.7. Figure \ref{angle_dist}(a) shows the results. The results indicated that the 
ASR of our adversarial clothing was higher than 
77\%in the range of 0-360 degrees. Besides, the ASR of the front (or back) of the adversarial 
clothing is better than ASR of the side, which may be due to the relatively small adversarial area on 
the side of the clothing.

Next, we analyzed the attack performance of adversarial clothing at different distances. The volunteers 
faced the camera (0 degrees) and the distance varied from 3 meters to 10 meters. We took 8 sample points 
between 3-10 meters and selected 100 images at each sample point. We also used ASR as the evaluation 
method and input these images into the YOLOv9 with the threshold of 0.7. 
The results are shown in Figure \ref{angle_dist}(b). The results indicated 
that our adversarial clothing had more than 78\% ASR in the range of 3-8 meters. When the distance was greater than 8 meters, the attack performance dropped quickly, because the adversarial pattern on the clothes was 
difficult to attack the detector in a smaller view.

\begin{figure}[tbp]
\centering
\includegraphics[width=1\columnwidth]{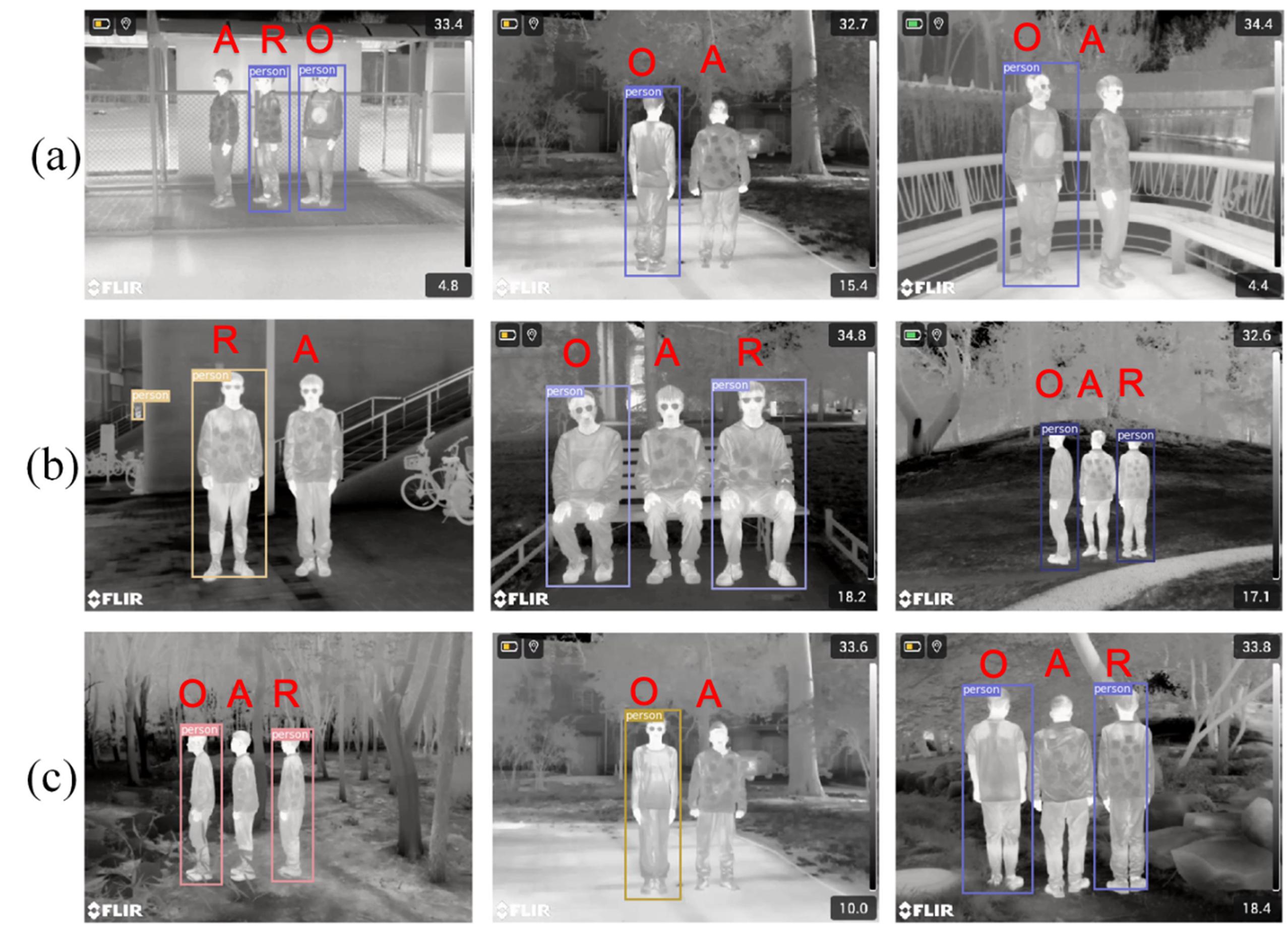} 
\caption{\added{Examples of different weather conditions: (a) sunny, (b) cloudy, and (c) foggy. A: adversarial clothing. R: random pattern clothing. O: ordinary clothing.} }
\label{weather}
\end{figure} 

\begin{table}[bp]
\centering
\caption{Evaluation across various weather conditions.}\label{tab_weather}
\begin{tabular}{cccc}
\toprule[1.1pt]  
Weather & Sunny & Cloudy & Foggy \\
\midrule  
ASR &  78.08\% &  75.83\% &  74.58\% \\
\bottomrule[1.1pt] 
\end{tabular}
\end{table}

\begin{table}[bp]
\centering
\caption{Evaluation across various landscape conditions.} \label{tab_landscape}
\begin{tabular}{cccc}
\toprule[1.1pt]  
Landscapes & City Road & Hillside & Riverside \\
\midrule  
ASR  &  76.25\% &  77.17\% &  75.08\% \\
\bottomrule[1.1pt] 
\end{tabular}
\end{table}

\subsubsection{Evaluation across various weather and landscape conditions}
\begin{figure}[tbp]
\centering
\includegraphics[width=1\columnwidth]{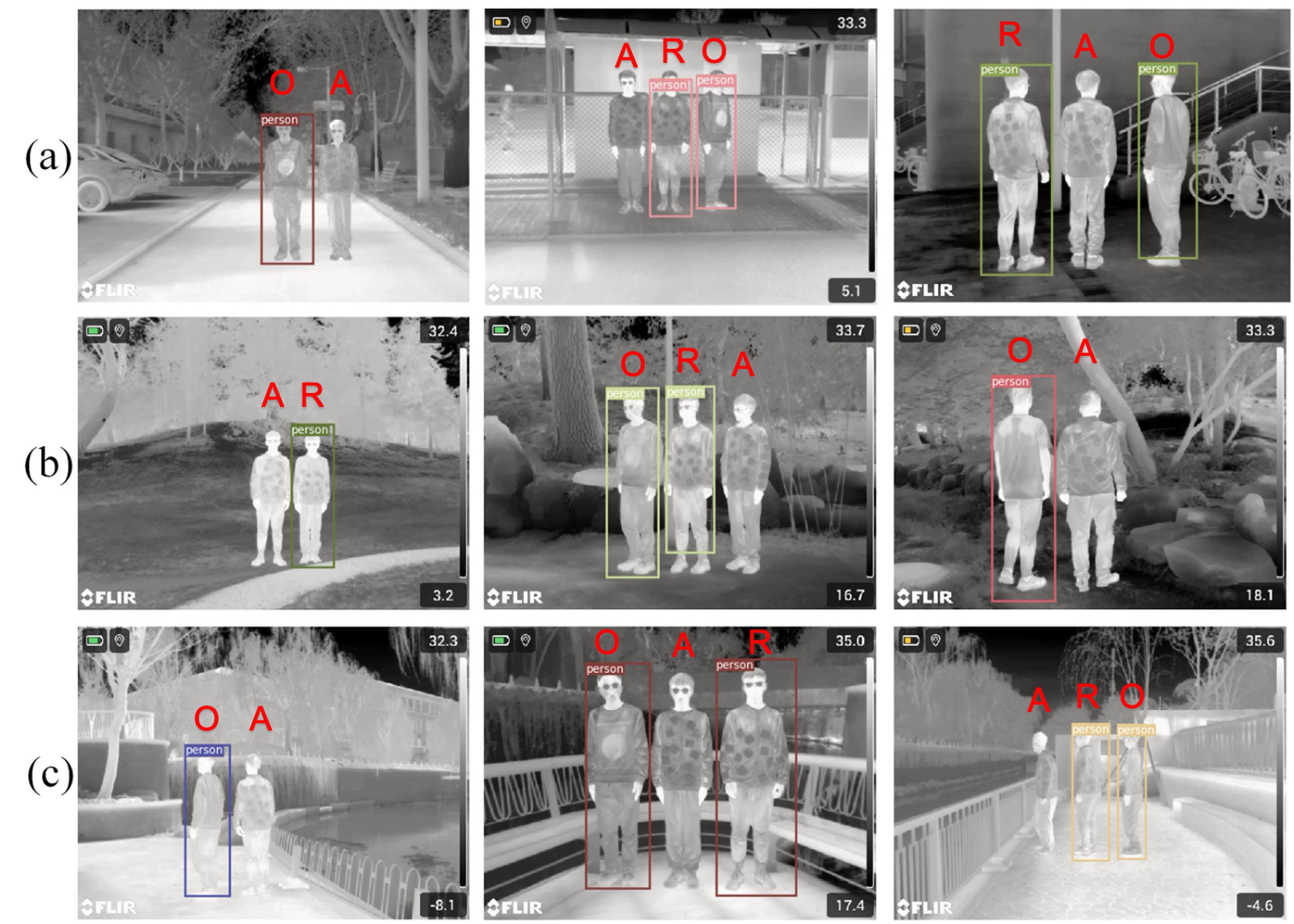} 
\caption{\added{Visual examples of different landscapes: (a) city road, (b) hillside and (c)riverside. A: adversarial clothing. R: random pattern clothing. O: ordinary clothing.} }
\label{landscape}
\end{figure}

We evaluated the performance of the adversarial clothing (Figure \ref{phy_cloth}(c)) under more diverse weather and landscape conditions. 
The weather conditions included sunny, cloudy, and foggy environments, while the landscapes included urban roads, hillsides, and riversides. 
For each case, we used a FLIR T560 infrared camera to capture 1200 images of volunteers wearing the adversarial clothing, 
and the attack success rate was computed accordingly. Other experimental settings followed those described in Section \ref{sec:physical attacks}.
The results are presented in Tables \ref{tab_weather} and \ref{tab_landscape}. \added{Several visual examples are shown in Figure \ref{weather} and \ref{landscape}.}

\begin{table*}[htbp]
\centering
\caption{Evaluation of the stability of adversarial clothing over time. }\label{tab_time}
\begin{tabular}{ccccccc}
\toprule[1.1pt]  
Time & 0h & 1h & 2h & 3h & 4h & 5h \\
\midrule  
ASR  &  83.17\% &  82.50\% &  82.00\% &  80.33\% &  78.67\% &  77.83\% \\
\bottomrule[1.1pt] 
\end{tabular}
\end{table*}

The results show that our method achieved an ASR over 74\% across various weather and landscape conditions, indicating its effectiveness. 
This performance can be attributed to the following factors. First, our algorithm pipeline incorporates simulations of 
multiple physical-world perturbations, such as random noise, temperature variations, diverse clothing texture changes, 
and random background switching. These augmentations enhance the robustness of our method to environmental variations in 
real-world scenarios. Second, thermal imaging inherently benefits from its penetration capability, allowing it to capture 
thermal images clearly even in the presence of environmental disturbances such as fog. As a result, the adversarial patterns 
we designed remain clear and effective in complex settings.

\subsubsection{Evaluation of the stability of adversarial clothing over time}
We evaluated the stability of the adversarial clothing over time. We recorded 3600 images of 
volunteers wearing the adversarial clothing (Figure \ref{phy_cloth}(c)) over a total duration of 5 hours. 
The other experimental settings were consistent with those described in Section \ref{sec:physical attacks}. 
The results are presented in Table \ref{tab_time}.

The results indicate that while the attack success rate of our adversarial clothing decreases slightly over time, 
it still maintains a success rate of 77.83\% after 5 hours, demonstrating the long-term stability of our method. 
This stability can be attributed to findings from our previous work\cite{zhu2022cvpr}, where we tested the thermal insulation properties 
of five materials, and found that the aerogel material had better thermal insulation properties than other materials and remained stable over 
time. Consequently, our aerogel-based clothing can consistently display the adversarial patterns, ensuring sustained attack effectiveness.

\begin{table*}[htbp]
\centering
\caption{Transferability in the digital world. The numbers are ASRs.}\label{tab_transfer}
\begin{tabular}{c|ccccccccc}
\toprule[1.1pt]  
\diagbox{Train}{Test} & DETR & DINO & DDQ & DAB & YOLOv7 & YOLOv8 & YOLOv11 & Mask & Cascade\\
\midrule  
YOLOv9& 29.75\% & 26.27\% & 27.51\% & 30.83\% & 53.96\% & 58.36\% & 50.28\% & 43.36\% & 40.06\% \\
Ensemble\_models& 81.03\% & 70.49\% & 75.20\%  & 73.42\% & 78.13\% & 80.28\% & 76.01\% & 72.57\%  & 70.35\% \\
\bottomrule[1.1pt] 
\end{tabular}

\end{table*}

\begin{table*}[htbp]
\centering
\caption{Transferability in the physical world. The numbers are ASRs.}\label{tab_phy_transfer}
\begin{tabular}{c|ccccccccc}
\toprule[1.1pt]  
\diagbox{Train}{Test} & DETR & DINO & DDQ & DAB & YOLOv7 & YOLOv8 & YOLOv11 & Mask & Cascade\\
\midrule  
YOLOv9& 21.69\% & 18.55\% & 19.84\% & 22.49\%  & 46.29\% & 50.11\% & 41.91\% & 39.22\%  & 31.06\% \\
Ensemble\_models& 70.05\% & 57.29\% & 60.85\%  & 56.25\% & 65.16\% & 68.33\% & 62.55\% & 68.04\%  & 59.73\% \\
\bottomrule[1.1pt] 
\end{tabular}

\end{table*}



\begin{figure}[bp]
\centering
\includegraphics[width=1\columnwidth]{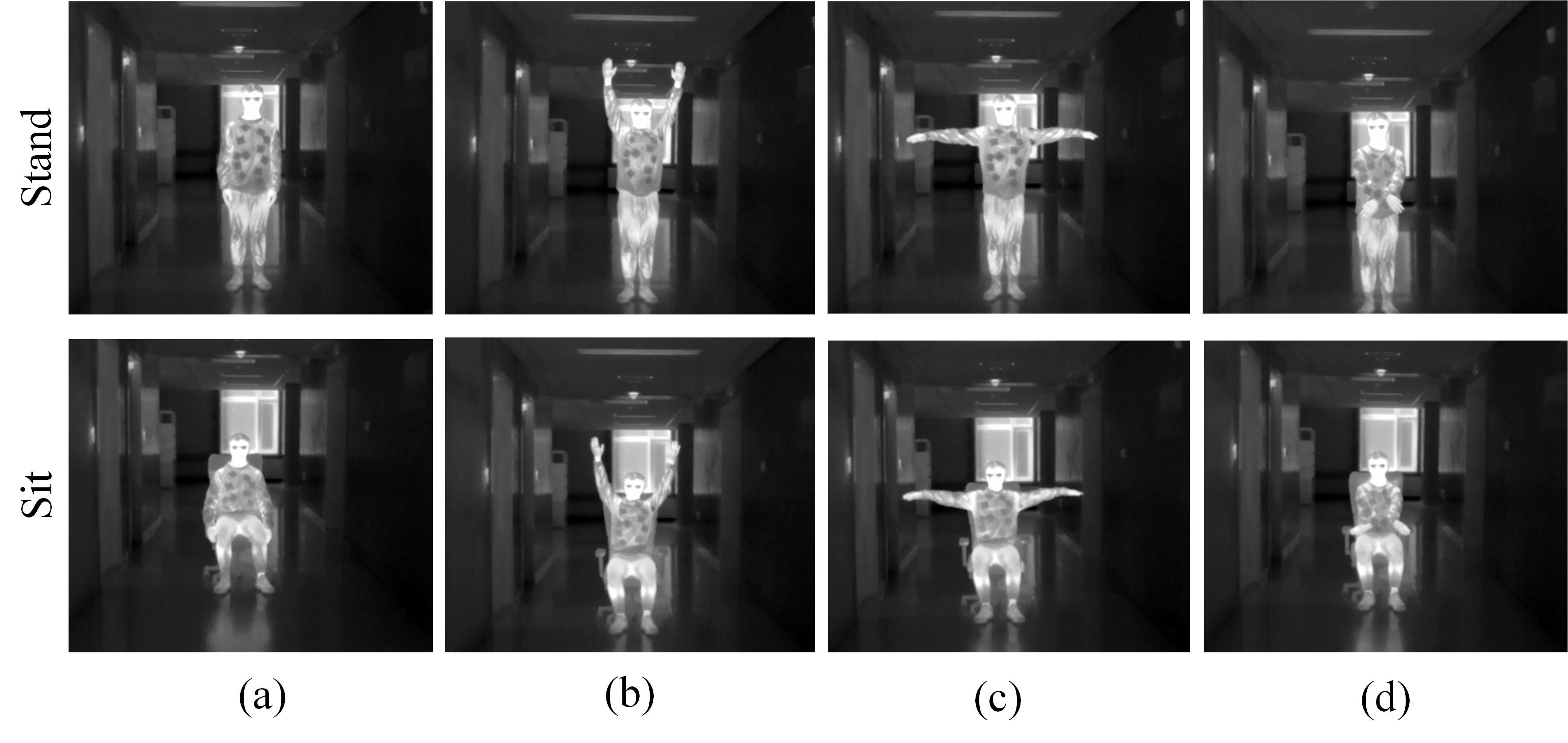} 
\caption{\added{Visual examples of volunteers with arms (a) hanging, (b) lifted, (c) spread, and (d) crossed in a standing (top row) or sitting (bottom row) status. } }
\label{posture}
\end{figure} 

\subsubsection{\added{Evaluation of the robustness against different postures}}
\added{We evaluated the robustness of our physical attack method against different postures. Volunteers could choose either 
a standing or sitting posture, and for each, their arm positions were categorized as arms hanging, arms lifted, 
arms spread, or arms crossed. Some visualization examples are shown in Figure \ref{posture}. For each case, we recorded six 
videos of approximately 10 seconds each and then computed the ASR. The results are shown in Table \ref{tab_stand} and Table \ref{tab_sit}. }

\added{The results indicate that, first, the ASR for standing poses is higher than for sitting poses, likely because our 
3D simulation model represents a standing posture. Second, the highest ASR is observed when the arms are hanging 
down, which aligns with our 3D model. The ASR decreases slightly (by 2.21-7.84\%) when the arms are lifted or 
spread, but it drops significantly  (20.27-21.32\%) when the arms are crossed. This is because our adversarial 
pattern design concentrates most of the patterns on the front or back of the body—regions that remain relatively 
stable across different poses and greatly influence the attack's effectiveness. While lifting or spreading the 
arms does not disturb these core areas, crossing the arms covers the adversarial patterns on the front, leading 
to a more pronounced decrease in ASR.}

\begin{table}[bp]
\centering
\caption{\added{Impact of arm posture on the ASR in a standing state. } }\label{tab_stand}
\begin{tabular}{ccccc}
\toprule[1.1pt]  
\added{Arm posture} & \added{Hanging}	 & \added{Lifted} & \added{Spread} & \added{Crossed} \\
\midrule  
\added{ASR}  &  \added{81.06\%}	 &  \added{76.85\%}	 &  \added{73.22\%}	 &  \added{60.79\%}  \\
\bottomrule[1.1pt] 
\end{tabular}
\end{table}

\begin{table}[bp]
\centering
\caption{\added{Impact of arm posture on the ASR in a sitting state.} }\label{tab_sit}
\begin{tabular}{ccccc}
\toprule[1.1pt]  
\added{Arm posture} & \added{Hanging}	 & \added{Lifted} & \added{Spread} & \added{Crossed} \\
\midrule  
\added{ASR}  &  \added{77.51\%}	 &  \added{75.30\%}	 &  \added{71.98\%}	 &  \added{56.19\%}  \\
\bottomrule[1.1pt] 
\end{tabular}
\end{table}

\subsection{Ensemble attacks}

We found that the attack transferability of a single model (YOLOv9) was limited. The ASR of optimized 
pattern (Figure \ref{v3_texture}(c)) against DETR\cite{carion2020end}, DINO\cite{zhang2022dino}, 
DDQ\cite{zhang2023dense}, DAB-DETR\cite{liu2022dabdetr}, YOLOv7\cite{wang2023yolov7}, YOLOv8\cite{sohan2024review}, YOLOv11\cite{khanam2024yolov11}, Mask-RCNN\cite{he2017mask},
and Cascade RCNN\cite{cai2019cascade} was only 29.75\%, 26.27\%, 
27.51\%, 30.83\% 53.96\%, 58.36\%, 50.28\%, 43.36\%, and 40.06\%, respectively in the digital world.

To improve the attack transferability, we used the model ensemble method as described in 
Section \ref{sec:3D model method}. We obtained a new 3D clothing pattern (Figure \ref{v3_texture}(d))
by integrating Deformable DETR\cite{zhu2020deformable}, YOLOv9, and Faster-RCNN \cite{journals/pami/RenHG017}. The optimization process was the same as 
Section \ref{sec:digital attacks}. As shown in Table \ref{tab_transfer}, the ASR of ensemble optimized pattern against 
DETR , DINO, DDQ, DAB-DETR, YOLOv7, YOLOv8, YOLOv11, Mask-RCNN, and Cascade RCNN was 81.03\%, 70.49\%, 75.20\%, 73.42\%, 78.13\%, 80.28\%, 76.01\%, 72.57\%, and 
70.35\%, respectively in the digital world. It indicated that the model ensemble effectively improved 
the attack transferability of the proposed method. 

Next, we manufactured physical clothes (Figure \ref{phy_cloth}(d)) based on the 3D clothing model 
obtained by ensemble model. We tested its attack performance in the physical world, and the results 
was shown in Table \ref{tab_phy_transfer}. 
The results showed that the ASRs of ensemble optimized pattern against DETR, DINO, DDQ, DAB-DETR, YOLOv7, YOLOv8, YOLOv11, Mask-RCNN,
and Cascade RCNN were obviously higher than that of the adversarial pattern optimized by single 
model YOLOv9 in the physical world. It indicated that our ensemble 
method could perform transferable attacks to unseen models in the real world.

The ensemble model-based method generally achieves better attack performance on black-box models compared to single-model 
optimization, possibly due to the following reasons: 

First, the ensemble model-based method can reduce the overfitting 
of the adversarial pattern to a single model, and the adversarial pattern obtained by the ensemble model can be more robust. 
Optimizing adversarial patterns for a single model (e.g., Model A) may yield patterns 
that are highly effective against Model A but fail on other models. In contrast, optimizing adversarial patterns simultaneously 
for multiple models (e.g., Models A, B, and C) requires the patterns to align with the decision boundaries of various models, 
making them more versatile in disrupting feature extraction across different models. As a result, the patterns are more likely 
to succeed against unseen models like Model D. 

Second, the ensemble model can capture the diversity of different models, 
and the adversarial pattern obtained by the ensemble model can attack the common features of different models.
Different models vary in architecture, training data, and loss functions. Integrating multiple models into the optimization process 
produces adversarial patterns that capture a broader range of disruptive modes, increasing resilience to inter-model differences. 
Such patterns are less reliant on exploiting specific weaknesses of a single model, resulting in higher attack success rates against previously unseen models.






\begin{table}[tbp] 
\caption{Evaluation of defense methods}\label{defense_methods}
\centering
\begin{tabular}{cc}
\toprule[1.1pt]  
Defense Methods & ASR decrease \\
\midrule  
SAC&  8.67\%  \\
\added{PAD} & \added{9.58\%} \\
\added{DIFFender} & \added{10.27\%} \\
Random Pattern Augmentation & 9.18\%  \\
Adversarial Training & 11.28\%\\
Pixel Mask & 7.31\%\\
Bit Squeezing & 3.85\% \\
Spatial Smoothing & 4.52\% \\
TVM & 6.19\% \\
\bottomrule[1.1pt] 
\end{tabular}
\end{table}

\subsection{Adversarial defense methods}
We tested seven typical adversarial defense methods to defend against the proposed 3D model-based attack 
in the digital world. These methods include SAC \cite{liu2022segment}, \added{PAD \cite{jing2024pad}, and DIFFender \cite{wei2024real},}  Random Patch Augmentation, Adversarial 
Training \cite{journals/corr/GoodfellowSS14}, Pixel Mask\cite{agarwal2021cognitive}, Bit Squeezing \cite{conf/ndss/Xu0Q18}, 
Spatial Smoothing \cite{conf/ndss/Xu0Q18}, and Total Variation Minimization (TVM) \cite{conf/iclr/GuoRCM18}.

For SAC, \added{PAD, and DIFFender,} we used the official code implementation of the original paper\cite{liu2022segment},\added{\cite{jing2024pad,wei2024real}}.
For Random Patch Augmentation, we incorporated the random black patches as data augmentation during 
pedestrian object detection training. The ratio of images with random black patches to clean images 
during the training process was 1:9 (Incorporating more adversarial examples in the training set 
such as the ratio of 2:8 or 3:7 causes the AP of clean images to drop by more than 10\%), which is 
to balance the robustness and AP of the model.
For Adversarial Training, we used data augmentation. We generated adversarial images using the 
methods described in Section \ref{sec:methods}. Then we added the adversarial images to the training 
set to train a more roboust model. The ratio of adversarial images to clean images during the training 
process was also 1:9. 
For Pixel Mask, we used one 20$\times$20 mask to erase the
adversarial clothing pattern at random positions.
For Bit Squeezing, we converted 8-bit images to 6-bit images. 
For Spatial Smoothing, we used the typical median filtering method, 
which was implemented in the SciPy \cite{scipy} Ndimage library.
For TVM, we used the TVM module in Adversarial Robustness Toolbox \cite{art2018} library.
The other optimization settings were the same as those described in Section \ref{sec:digital attacks}.

We used the adversarial pattern (Figure \ref{v3_texture}) to test
the effect of these defense methods. Results in Table \ref{defense_methods} indicated 
that the effect of these defense methods was limited. The most effective 
method (adversarial training)  decreased the ASR of YOLOv9 by only
11.28\% (from 86.38\% to 75.10\%), while the proposed attack method still achieved the ASR of 75.10\%.

\section{Conclusion}
This paper presents a new infrared physical attack method based on 3D modeling. We constructed 3D human and clothing models under infrared imaging, and simulated their attack effects in different scenes.
To simulate the infrared characteristics more realistically, we propose a method to build infrared 
3D models with real infrared photos. We also build different texture maps for 3D clothing model to 
simulate changeable infrared clothing characteristics in different times and places.
We designed a pattern of 3D infrared clothing by optimizing the positions and orientations of black 
patches, and manufactured a physical clothing based on the thermal-insulated material aerogel. 
Our clothing looks like ordinary clothing in the 
visible light view, but can hide from infrared detectors at multiple angles in different scenes in the 
physical world. We effectively improved the attack transferability to unknown models by using the model 
ensemble technique.

\section{Limitation}
Our method also has several limitations. First, when the adversarial clothes are too far from the camera, it is difficult to attack the 
infrared detectors in a quite small view. This is reasonable because the adversarial images 
are blurry when scaled to very small sizes.
Second, when the physical adversarial pattern is severely occluded, its attack effect decreases 
a lot. This is because the occluded adversarial pattern in the physical world deviates greatly from the adversarial pattern 
optimized in the digital world. We will address these challenges in future work.

\ifCLASSOPTIONcompsoc
  \section*{Acknowledgments}
\else
  \section*{Acknowledgment}
\fi

This work was supported by the National Natural Science Foundation of China (No. U2341228).

\ifCLASSOPTIONcaptionsoff
  \newpage
\fi

\begin{IEEEbiography}[{\includegraphics[width=1in,height=1.25in,clip,keepaspectratio]{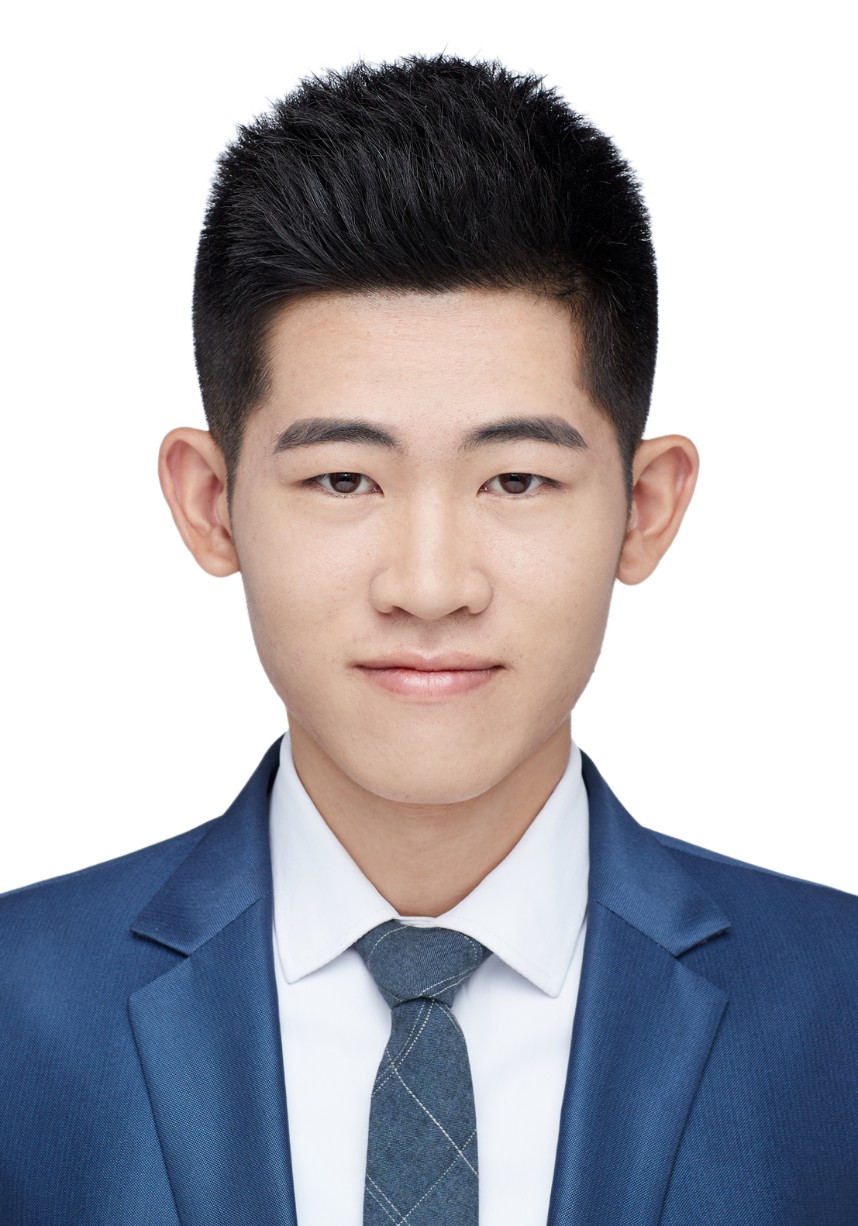}}]{Xiaopei Zhu}
received his Ph.D. degree in Tsinghua University, Beijing, 
China, in 2024. He is now a postdoctoral fellow in Tsinghua University, working with Prof. Jun Zhu. 
His current research interests include computer vision, adversarial example, generative model, 3D modeling, AI4sicence, etc. 
He has published several papers in CVPR, AAAI, ACM MM, etc. 
\end{IEEEbiography}


\begin{IEEEbiography}[{\includegraphics[width=1in,height=1.25in,clip,keepaspectratio]{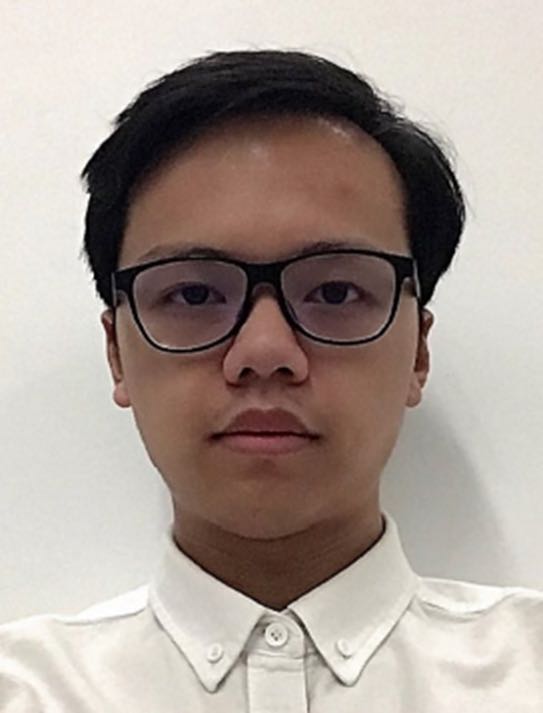}}]{Siyuan Huang}
received a B.E. degree in Software Engineering from the Wuhan University of Technology, Wuhan, China, in 2018, 
and M.S. in Computer Science from the George Washington University, Washington, DC, US, in 2020. He is currently a 
Ph.D. student at the Department of Electrical and Computing Engineering, Johns Hopkins University, Baltimore, US. 
His current research interest is computer vision. Previously he was a Graduate Research Assistant at Tsinghua 
Laboratory of Brain and Intelligence, Tsinghua University, and an Associate Researcher at Alternative Computing Group, 
National Institute of Standards and Technology.
\end{IEEEbiography}

\begin{IEEEbiography}[{\includegraphics[width=1in,height=1.25in,clip,keepaspectratio]{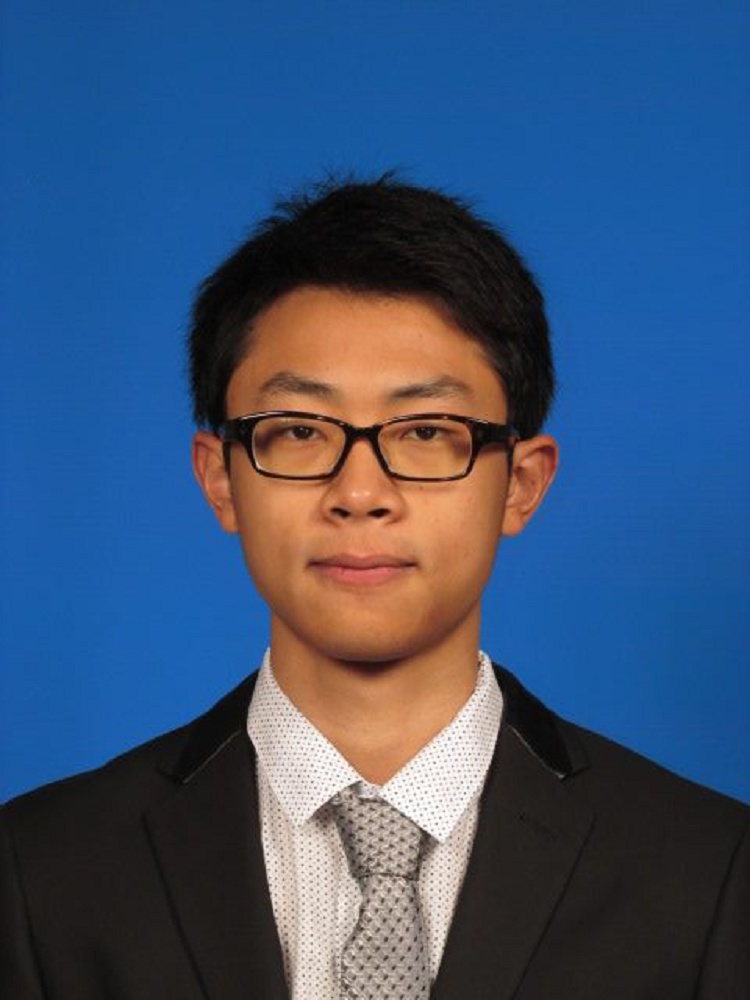}}]{Zhanhao Hu}
is a postdoc in the Department of Electrical Engineering and Computer Sciences at the University of California, Berkeley. 
He received his Ph.D. in Computer Science and Technology from Tsinghua University in 2023 and his B.S. in Mathematics and 
Physics from Tsinghua University in 2017. His current research includes security issues in deep learning, especially in 
Computer Vision and Large Language Models.
\end{IEEEbiography}


\begin{IEEEbiography}[{\includegraphics[width=1in,height=1.25in,clip,keepaspectratio]{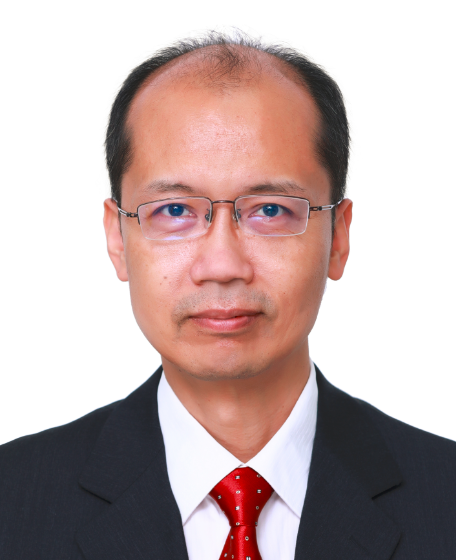}}]{Jianmin Li}
received the Ph.D. degree in computer
applications from the Department of Computer
Science and Technology, Tsinghua University,
in 2003. He is currently an Associate
Professor with the Department of Computer Science
and Technology, Tsinghua University. His
main research interests include image and video
analysis, image and video retrieval, and machine
learning. He has published over 80 journal and
conference papers.
\end{IEEEbiography}




\begin{IEEEbiography}[{\includegraphics[width=1in,height=1.25in,clip,keepaspectratio]{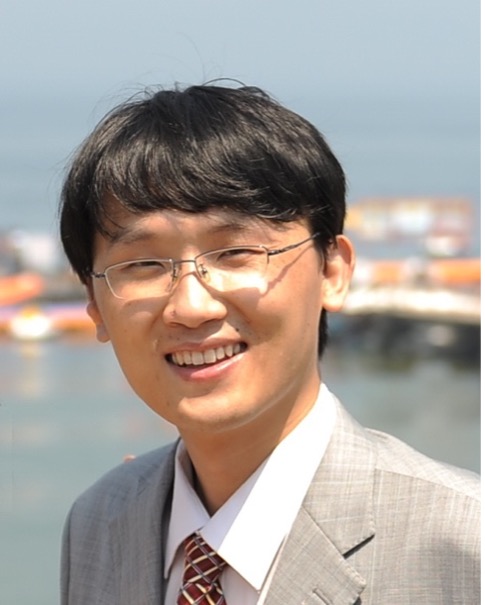}}]{Jun Zhu}
  (Fellow, IEEE)
  received the BS and PhD
  degrees from the Department of Computer Science
  and Technology, Tsinghua University, where he is
  currently a Bosch AI professor. He was a postdoctoral
  fellow and adjunct faculty with the Machine
  Learning Department, Carnegie Mellon University.
  His research interest is primarily on developing machine
  learning methods to understand scientific and
  engineering data arising from various fields. He regularly
  serves as senior area chairs and area chairs with
  prestigious conferences, including ICML, NeurIPS,
  ICLR, IJCAI and AAAI. He was selected as ``AI's 10 to Watch'' by IEEE
  Intelligent Systems. He is a fellow of AAAI, and an associate editor-in-chief
  of IEEE Transactions on Pattern Analysis and Machine Intelligence.
\end{IEEEbiography}

\begin{IEEEbiography}[{\includegraphics[width=1in,height=1.25in,clip,keepaspectratio]{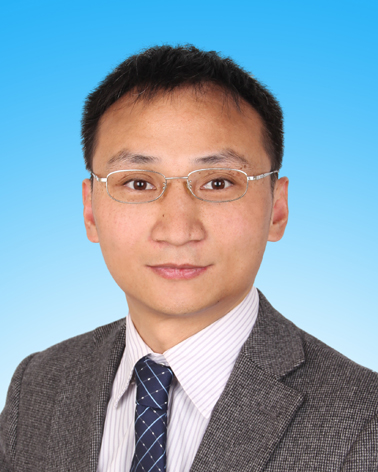}}]{Xiaolin Hu}
(S'01, M'08, SM'13) received B.E. and M.E. degrees in automotive engineering from the Wuhan University of 
Technology, Wuhan, China, in 2001 and 2004, respectively, and a Ph.D. degree in automation and computer-aided 
engineering from the Chinese University of Hong Kong, Hong Kong, in 2007. 
He is currently an Associate Professor at the Department of Computer Science and Technology, 
Tsinghua University, Beijing, China. His current research
interests include deep learning and computational
neuroscience.  At present, he is an
Associate Editor of IEEE Transactions on Pattern Analysis and Machine Intelligence (TPAMI), 
IEEE Transactions on Image Processing (TIP) and Cognitive Neurodynamics.
Previously he was an Associate Editor of the IEEE Transactions on Neural Networks and Learning Systems.
\end{IEEEbiography}
\end{document}